\documentclass{article}
\usepackage{iclr2025_conference,times}
\usepackage{hyperref}
\usepackage{url}

\usepackage[T1]{fontenc}
\usepackage[utf8]{inputenc}
\usepackage{microtype}
\usepackage{inconsolata}
\usepackage{graphicx}
\usepackage{tcolorbox}
\usepackage{amsmath}
\usepackage{booktabs}
\usepackage{multirow}
\usepackage{colortbl}
\usepackage{subcaption}
\usepackage{listings}
\usepackage{subcaption}
\usepackage{booktabs}
\usepackage{tabularx} 
\usepackage{wrapfig} 
\usepackage{cleveref}
\usepackage{booktabs}
\usepackage{siunitx}
\usepackage{colortbl}
\usepackage{caption}
\usepackage{subcaption}
\usepackage{booktabs}
\usepackage{graphicx}
\usepackage{enumitem}
\usepackage{array}

\definecolor{darkblue}{rgb}{0.1,0.1,0.44}
\definecolor{darkred}{rgb}{0.55,0.0,0.0}
\definecolor{gray}{rgb}{0.5,0.5,0.5}

\title{Turning Up the Heat: Min-\(p\) Sampling for \\ Creative and Coherent LLM Outputs}

\author{
  Nguyen Nhat Minh\thanks{Equal Contribution. Correspondence to Nguyen Nhat Minh at \texttt{minh1228@gmail.com}.}$^{*1}$,
  Andrew Baker$^{*2}$,
  Clement Neo$^{*1}$, \\
  \textbf{Allen Roush}$^{3}$,\textbf{ Andreas Kirsch}$^{2}$,
  \textbf{Ravid Shwartz-Ziv}$^{3,4}$ \\
  $^{1}$Apart Research \quad
  $^{2}$Independent \quad
  $^{3}$Wand.ai \quad
  $^{4}$New York University
  \\
}

\iclrfinalcopy

\begin{document}

\maketitle

\begin{abstract}
Large Language Models (LLMs) generate text by sampling the next token from a probability distribution over the vocabulary at each decoding step. Popular sampling methods like top-\(p\) (nucleus sampling) often struggle to balance quality and diversity, especially at higher temperatures which lead to incoherent or repetitive outputs. We propose \textbf{min-\(p\) sampling}, a dynamic truncation method that adjusts the sampling threshold based on the model's confidence by using the top token's probability as a scaling factor. Our experiments on benchmarks including GPQA, GSM8K, and AlpacaEval Creative Writing show that min-\(p\) sampling improves both the quality and diversity of generated text across different model families (Mistral and Llama 3) and model sizes (1B to 123B parameters), especially at higher temperatures. Human evaluations show a preference for min-\(p\), in both text quality and creativity. Min-\(p\) sampling has been adopted by popular open-source LLM frameworks, including Hugging Face Transformers, VLLM, and many others, highlighting its considerable impact on improving text generation quality.
\end{abstract}

\section{Introduction}

A central challenge for Large Language Models (LLMs) is the trade-off between creativity and coherence during text generation. Popular sampling strategies like top-$p$ sampling \citep{holtzman2020topp} and temperature scaling \citep{ackley1985learning} are widely adopted, but increasing temperature to enhance diversity often reduces coherence, while more conservative sampling limits creativity.

We address the \textit{quality-diversity tradeoff} with \textbf{min-$p$ sampling}, a dynamic truncation method that adjusts the sampling threshold based on model confidence using the top token's probability as a scaling factor. This approach includes diverse options when the model is uncertain and keeps only high-confidence tokens when confident, balancing creativity and coherence even at high temperature.

Our experiments on benchmarks including \textbf{GPQA} \citep{rein2023gpqa}, \textbf{GSM8K} \citep{cobbe2021training}, and \textbf{AlpacaEval Creative Writing} \citep{alpaca_eval}. Our results show that min-\(p\) sampling outperforms top-\(p\) and other popular decoding methods such as top-\( k \) \citep{fan-etal-2018-hierarchical} and $\eta$-sampling \citep{hewitt-etal-2022-truncation} by maintaining coherence while allowing for increased diversity, particularly at high-temperatures. We also conducted a comprehensive \textbf{human evaluation} to compare the quality and diversity of text generated by min-\(p\) with that generated by traditional sampling methods. The results indicate a clear preference for min-\(p\), with participants rating min-\(p\) outputs as superior in quality and diversity.

The contributions of this paper are as follows:

\begin{itemize}
    \item We introduce \textbf{min-\(p\) sampling}, a novel dynamic truncation method that effectively balances creativity and coherence in LLM-generated text, particularly at high temperatures.
    \item We present comprehensive \textbf{experimental results} across benchmarks, model families and sizes, demonstrating that in practice min-\(p\) consistently improves quality and diversity of generated text, especially at higher temperatures, compared to sampling methods like top-\(p\).
    \item We validate the practical utility of \textbf{min-\(p\)} through an extensive \textbf{human evaluation}, showing instances where human evaluators prefer min-\(p\) outputs over those generated by other methods in terms of both quality and diversity.
    \item We provide  practical \textbf{empirical guidelines} for hyperparameter selection and usecases.
    \item The rapid \textbf{adoption of min-\(p\)} by the open-source LLM community highlights its practical benefits, even for large reasoning models such as Deepseek R1 \citep{deepseekai2025deepseekr1incentivizingreasoningcapability}.
\end{itemize}

By introducing min-\(p\) and offering empirical guidelines for its use, we enable exploration of high-temperature creative text generation without sacrificing coherence. Our results demonstrate that min-\(p\) is a viable and superior alternative to existing sampling methods, both at standard and high-temperature settings, respectively, marking a major contribution to generative language modeling.

\section{Related Work}
\label{sec:related_work}

We review current sampling methods and quality-diversity tradeoff, establishing motivation for min-\(p\).

\paragraph{Greedy Decoding and Beam Search.} 
These deterministic decoding strategies select the highest probability token at each step \citep{freitag2017beam}, which often produces repetitive and generic text due to lack of diversity. Beam search also incurs a notable runtime performance penalty.

\paragraph{Stochastic Sampling Methods.} These methods inject diversity via randomness in token selection. \textbf{Temperature scaling} adjusts distribution sharpness to balance diversity and coherence \citep{ackley1985learning}; however, higher temperatures often give incoherent outputs, limiting applicability. \textbf{Top-$k$ sampling} selects from the $k$ most probable tokens \citep{fan-etal-2018-hierarchical}, offering a simple way to prevent sampling unlikely tokens, but doesn't dynamically adapt to varying confidence levels across contexts. 

\textbf{Top-$p$ sampling} (nucleus sampling) restricts the token pool to those whose cumulative probability exceeds a total threshold $p$ \citep{holtzman2020topp}, balancing quality and diversity by preserving the "nucleus" of high-probability tokens. However, at higher temperatures, top-$p$ can still allow low-probability tokens into the sampling pool, leading to incoherent outputs. This specific trade-off between creativity and coherence at high temperatures is a key limitation we address with min-\(p\).

\paragraph{Entropy-Based Sampling.} Recent approaches like \textbf{($\eta$-sampling)} and \textbf{mirostat sampling} dynamically adjust the sampling pool based on token distribution entropy \citep{hewitt-etal-2022-truncation, basu2021mirostat}. While these show promise, they often require complex parameter tuning and can be computationally expensive. We detail our experimental challenges running $\eta$ sampling in \Cref{app:etaisslow}.

\paragraph{Dynamic Threshold Methods.} Recent work has explored adaptive threshold mechanisms that adjust based on model confidence. \citet{li2023contrastivedecodingopenendedtext} introduced an ``adaptive plausibility constraint'' in their contrastive decoding framework, which filters tokens using a threshold proportional to the maximum probability: $\alpha \cdot \max_w P(w|x_{<i})$. This constraint is similar to min-\(p\) as a threshold relative to model confidence rather than absolute. The key distinction is that \citet{li2023contrastivedecodingopenendedtext} developed this as a constraint within a contrastive objective using expert and amateur models, min-\(p\) treats confidence-scaled truncation as the standalone sampling mechanism, a much simpler approach.

\section{Min-\texorpdfstring{$p$}{p} Sampling}
\label{sec:min_p_sampling}

The core idea of \textbf{min-$p$ sampling} is to dynamically adjust the sampling threshold based on the model's confidence at each decoding step, improving sensitivity to context and uncertainty while better balancing creativity and coherence, especially at high temperatures.

\subsection{Overview of Min-\texorpdfstring{$p$}{p} Sampling}

In autoregressive generation, a language model predicts the probability distribution over the vocabulary for the next token, conditioned on the existing sequence. At each step, the model selects a token from this distribution. Min-\(p\) sampling is a stochastic method that varies its truncation threshold based on the model's confidence, making the threshold context-sensitive.

Formally, at each time step $t$, let $\mathcal{V}$ denote the vocabulary, and $P(x_t \mid x_{1:t-1})$ represent the conditional probability distribution over the vocabulary for the next token $x_t$. Min-\(p\) sampling works as follows:

\begin{enumerate}
    \item \textbf{Calculate the Maximum Probability:} Identify the maximum probability token in the distribution, denoted as $p_{\max} = \max_{v \in \mathcal{V}} P(v \mid x_{1:t-1})$.
    \item \textbf{Truncation Threshold:} Set a base probability threshold, $p_{\text{base}} \in (0, 1]$ Scale it by $p_{\max}$:
    \begin{equation}
        p_{\text{scaled}} = p_{\text{base}} \times p_{\max}
    \end{equation}
    This threshold ensures that tokens with sufficiently high relative probabilities are considered while filtering out less probable tokens in a context-dependent manner.
    \item \textbf{Define Sampling Pool:} Construct pool $\mathcal{V}_{\text{min}}$ of tokens with probabilities $\geq p_{\text{scaled}}$:
    \begin{equation}
        \mathcal{V}_{\text{min}} = \{v \in \mathcal{V} : P(v \mid x_{1:t-1}) \geq p_{\text{scaled}} \}
    \end{equation}
    \item \textbf{Sample from Pool:} Sample the next token $x_t$ from the normalized set $\mathcal{V}_{\text{min}}$:
    \begin{equation}
        P'(v) = \frac{P(v \mid x_{1:t-1})}{\sum_{v' \in \mathcal{V}_{\text{min}}} P(v' \mid x_{1:t-1})} \quad \text{for } v \in \mathcal{V}_{\text{min}}
    \end{equation}
\end{enumerate}

\subsection{Intuition Behind Min-\texorpdfstring{$p$}{p} Sampling}

The key intuition behind min-\(p\) sampling is that \textbf{token truncation thresholds are relative and depend on how certain the distribution is for that token}, and not absolute thresholds. When the model is highly confident about the next token (i.e., \( p_{\max} \) is high), min-\(p\) restricts the pool to high-probability candidates to maintain coherence. Conversely, when the model is less confident, relaxing the sampling pool allows for a more creative and diverse generation. Unlike top-\(p\) sampling, which truncates the distribution based on a fixed cumulative probability, min-\(p\) dynamically adjusts the threshold based on the model's confidence, leading to more context-sensitive generation.

\begin{figure}[t]
    \centering
    \begin{subfigure}[b]{0.48\textwidth}
        \centering
        \includegraphics[width=\textwidth]{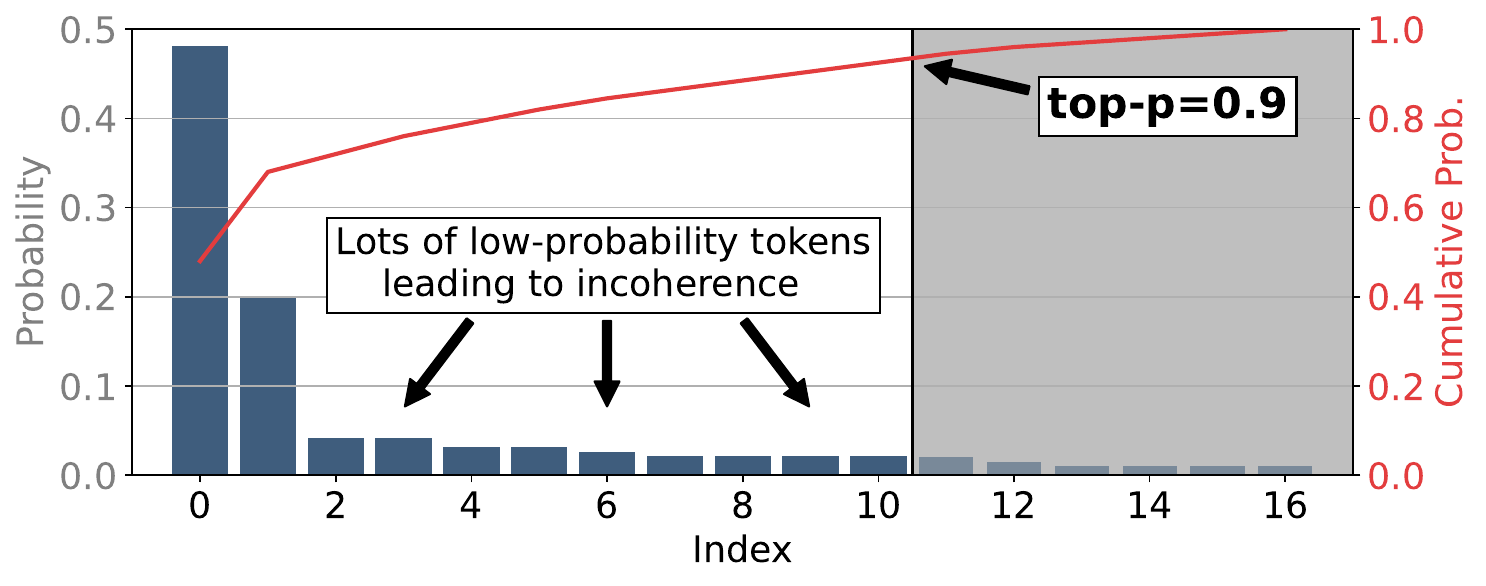}
        \caption{}
        \label{fig:subfig1}
    \end{subfigure}
    \hfill
    \begin{subfigure}[b]{0.48\textwidth}
        \centering
        \includegraphics[width=\textwidth]{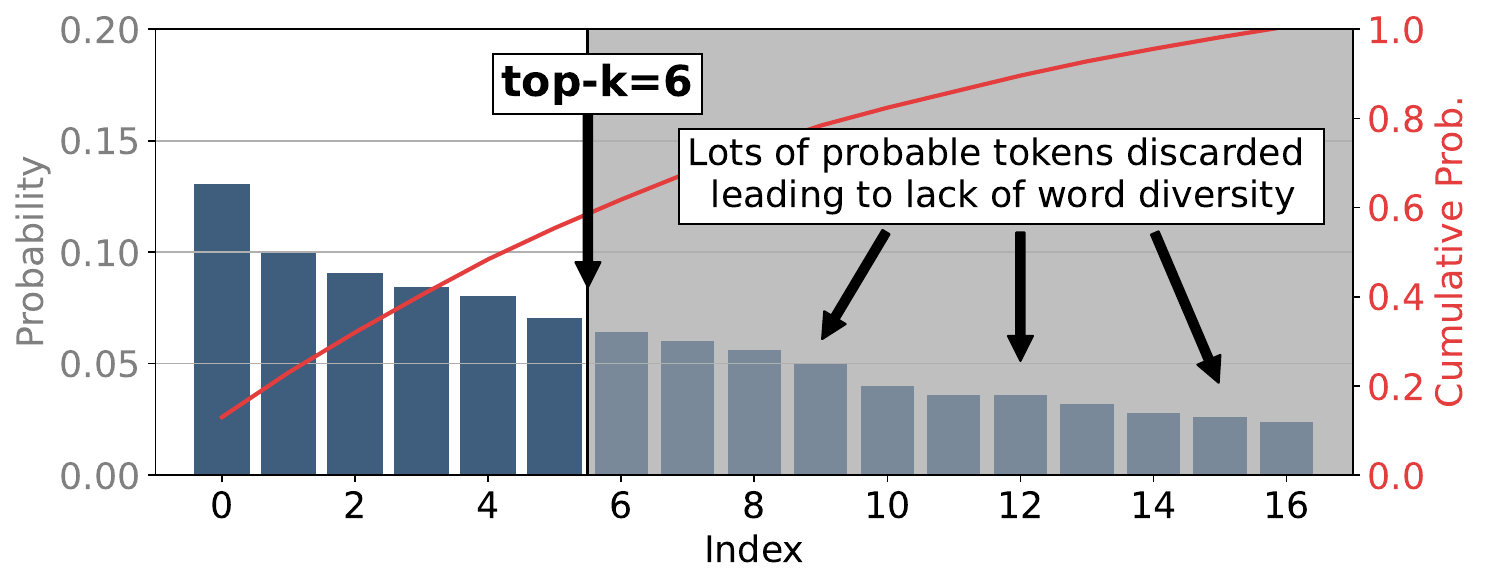}
        \caption{}
        \label{fig:subfig2}
    \end{subfigure}
    
    \vspace{1em}
    
    \begin{subfigure}[b]{0.48\textwidth}
        \centering
        \includegraphics[width=\textwidth]{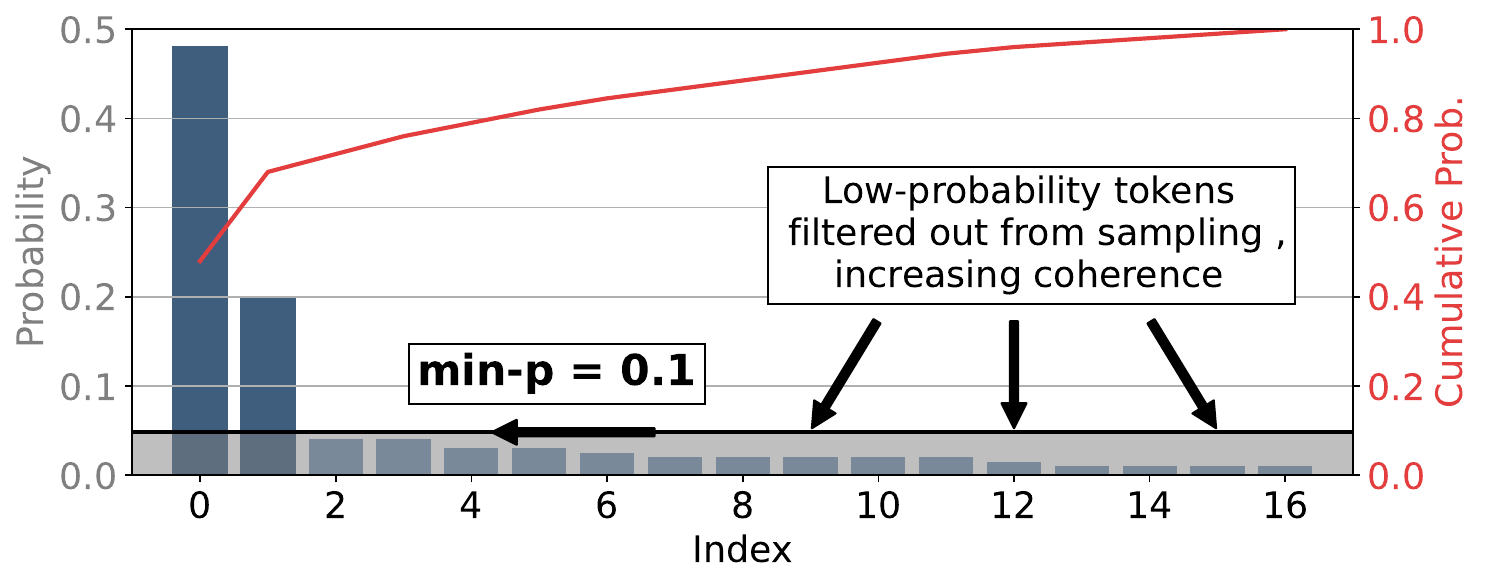}
        \caption{}
        \label{fig:subfig3}
    \end{subfigure}
    \hfill
    \begin{subfigure}[b]{0.48\textwidth}
        \centering
        \includegraphics[width=\textwidth]{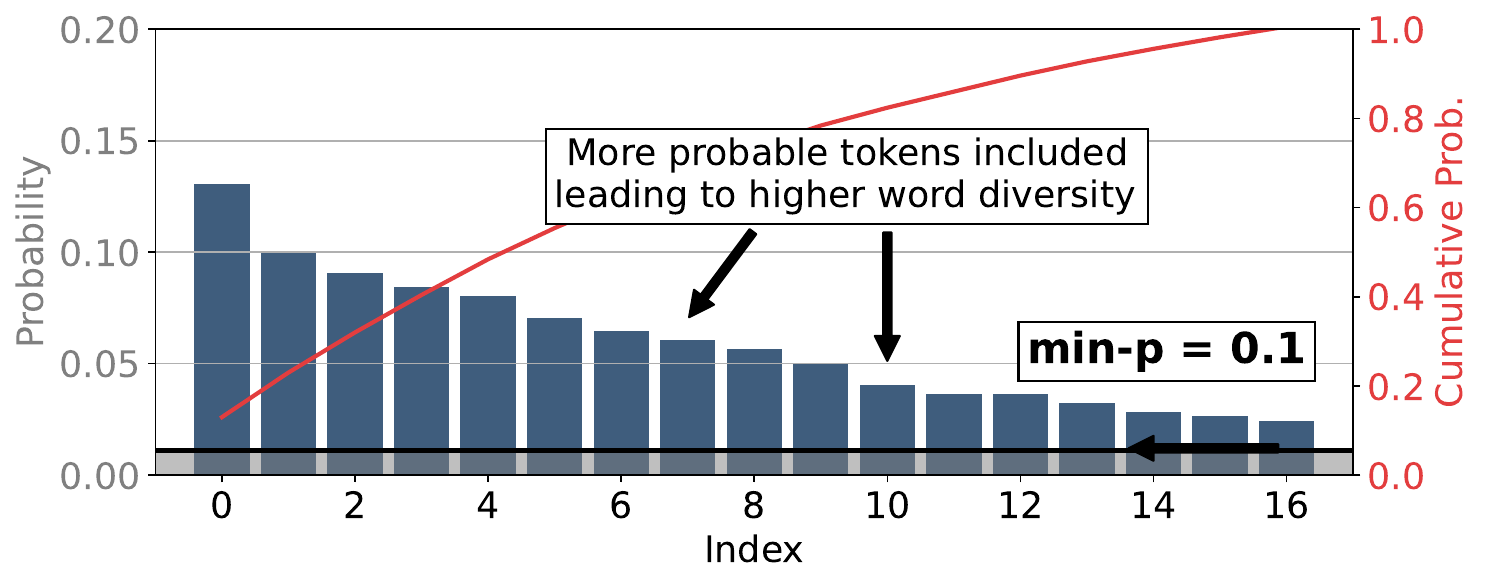}
        \caption{}
        \label{fig:subfig4}
    \end{subfigure}
    
   \caption{Comparison of sampling methods. (a) \textbf{Top-\(p\) sampling} in a high-certainty case. (b) \textbf{Top-$k$ sampling} in a low-certainty case. (c) \textbf{Min-\(p\) sampling} in a high-certainty case. (d) \textbf{Min-\(p\) sampling} in a low-certainty case. Min-\(p\) sampling dynamically adjusts its threshold based on the model's confidence, focusing on likeliest tokens when confident and allowing more diverse options when uncertain. This dynamic threshold balances coherence and diversity better than top-\(p\) and top-$k$.}

    \label{fig:sampling}
\end{figure}

Figure \ref{fig:sampling} illustrates the effects of different sampling methods, including min-\(p\), on token probability distributions. In subfigure (a), we show an initial probability distribution over tokens. Subfigures (b), (c), and (d) demonstrate how top-\(p\), top-$k$, and min-\(p\) sampling truncate this distribution. Min-\(p\) dynamically adjusts its filtering threshold based on the model's confidence, focusing on high-probability tokens when confident and including more diverse options when uncertain. This dynamic behavior helps min-\(p\) balance coherence and diversity more effectively than top-\(p\) and top-$k$ sampling.

\subsection{Advantages Over Existing Methods}

Min-$p$ sampling provides several advantages over existing sampling methods:

\paragraph{Balancing Creativity and Coherence.}
Unlike more static thresholds like top-$p$ and top-$k$ that may allow either overly diverse (incoherent) or conservative (repetitive) token choices in the token sampling pool, min-$p$'s can dynamically adjust its sampling threshold to different contexts within the same generated sequence based on the model's confidence in its outputs.

\paragraph{Robustness at High Temperatures.}
A key limitation of existing sampling methods is their performance at high temperatures. As the temperature increases, token probabilities become more uniform, allowing unlikely tokens to be selected, resulting in incoherent text. Min-\(p\) addresses this by scaling the truncation threshold proportional to the model's confidence, preserving output coherence even at higher temperatures. This makes it particularly valuable for tasks that benefit from high creativity, such as storytelling or generating diverse solutions \citep{tajwar2025traininggenerallycuriousagent}.

\paragraph{Simplicity.}
Min-\(p\) is simple, requiring minimal calculations. Unlike methods involving auxiliary models \citep{li2023contrastivedecodingopenendedtext} or complex entropy-based adjustments \citep{hewitt-etal-2022-truncation}, min-\(p\) integrates easily into existing LLM inference pipelines without substantial overhead, offering practical advantages for both research and real-world applications compared to entropy-based methods such as \(\epsilon\) and \(\eta\) sampling \citep{hewitt2022truncation}, as we discuss in \Cref{app:etaisslow}.

\subsection{Implementation Details}

Min-$p$ requires minimal changes to standard LLM decoding pipelines. Here, we provide reference:

\paragraph{Integration into Decoding Pipelines.} Min-\(p\) sampling has been implemented as a logits processor in frameworks like Hugging Face Transformers \citep{wolf-etal-2020-transformers} and , vLLM \citep{kwon2023efficient}. After applying temperature scaling, the scaled threshold \( p_{\text{scaled}} \) is computed, and tokens with probabilities below this threshold are filtered out before sampling. These operations are efficiently implemented using vectorized computations, adding negligible overhead to the decoding process.

\paragraph{Parameter Selection Guidelines.}
\begin{itemize}
    \item \textbf{Choosing the Base Threshold (\( p_{\text{base}} \)):} Setting \( p_{\text{base}} \) between 0.05 to 0.1 effectively balances creativity and coherence across tasks and models. Higher values of \( p_{\text{base}} \) (e.g., 0.5-0.7 or even close to 1) can help maintain coherence and improve benchmark scores at very high temperatures, typically at the cost of diversity (see \Cref{tab:performance_results}).
    \item \textbf{Temperature Settings:} Min-\(p\) works well across a wide range of temperatures, with higher temperatures (e.g., \( \tau = 2 \) or \( \tau = 3 \)) enhancing diversity with much lower loss of coherence.
    \item \textbf{Combining with Other Techniques:} While min-\(p\) sampling can work with other sampling methods or repetition penalties, we recommend using it as the primary truncation method to fully leverage its dynamic capabilities. Our experiments with combined samplers (see \Cref{tab:benchmark_comparison}) show that multiple truncation methods can lead to double normalization issues and suboptimal performance, as each method's hyperparameters have so far been optimised for standalone use. We discuss potential research into combined sampling in \Cref{sec:limitations}.
\end{itemize}

\paragraph{Availability of Resources} To facilitate adoption, reference implementations of min-\(p\) are available:

\begin{itemize}
    \item \textbf{Community Adoption:} Min-$p$ sampling has been integrated in widely-used open-source frameworks such as Hugging Face Transformers, vLLM, SGLang and many others repositories \citep{zheng2024sglangefficientexecutionstructured}, which collectively have accrued over 667,000 GitHub stars. This integration, coupled with extensive downstream usage reflected in dependents of these frameworks (e.g., over 290,000 dependent repositories for Transformers alone), underscores the method's availability and potential for impact (see Appendix~\ref{sec:adoption-stats} for detailed statistics). While framework popularity reflects many factors, the rapid adoption of min-\(p\) demonstrates its perceived utility by developers and researchers, and its availability to the community.
    \item \textbf{Project Repository:} Reference code implementation is available at our project repository.\footnote{\url{https://github.com/menhguin/minp_paper}}
\end{itemize}

\paragraph{Relevance to frontier reasoning models.}
Further validating min-\(p\)'s utility, Unsloth, a major inference provider whose DeepSeek R1 \citep{deepseekai2025deepseekr1incentivizingreasoningcapability} implementation has received > as dy 1 shows min-\(p\) increasing token diversity, while Case Study 2 demonstrates min-\(p\) better preserving coherence in high-confidence scenarios.

\section{Case Studies: Illustrative Examples}
\label{sec:case_studies}

To provide qualitative insights into how \textbf{min-\(p\) sampling} compares to existing methods, we present two case studies that highlight differences in token truncation, especially at higher temperatures. These examples illustrate min-$p$'s dynamic behavior of min-\(p\) in practice. This visualization was originally created by open-source LLM hobbyist \citet{dalmaso2024llmeval}, and reproduced in this paper.

\begin{table}[t]
    \centering
    \caption{\textbf{Token probability comparison between top-\(p\) and min-\(p\).} Case Study 1 shows min-\(p\) increasing token diversity. Case Study 2 shows min-\(p\) staying coherent in high-confidence scenarios.}
    \label{tab:token_probs_case_studies}
    \small 
    \setlength{\tabcolsep}{3pt} 
    \begin{tabular}{p{0.48\textwidth} p{0.48\textwidth}}
    \begin{minipage}[t]{\linewidth}
        \centering
        \subcaption{Case Study 1: Low-Certainty Next Token}
        \fcolorbox{black}{gray!10}{
        \begin{minipage}{0.95\linewidth}
            \textbf{Prompt:} \emph{"You will pay for what you have done," she hissed, her blade flashing in the moonlight. The battle that ensued \_\_\_\_}
        \end{minipage}
        }
        \vspace{0.2cm}
        \rowcolors{2}{gray!10}{white}
        \begin{tabular}{p{1.8cm} *{4}{>{\centering\arraybackslash}p{0.9cm}}}
            \toprule
            \textbf{Token} & \textbf{\( \tau{=}1 \)} & \textbf{\( \tau{=}3 \)} & \textbf{Top-\(p\)} & \textbf{Min-\(p\)} \\
            \midrule
            was     & 70.3 & 11.9 & 13.1 & 18.5 \\
            lasted  & 9.5  & 6.1  & 6.7  & 9.5 \\
            between & 6.2  & 5.3  & 5.9  & 8.2 \\
            left    & 4.5  & 4.8  & 5.3  & 7.4 \\
            would   & 3.2  & 4.3  & 4.7  & 6.6 \\
            seemed  & 0.5  & 2.3  & 2.5  & 3.5 \\
            \bottomrule
        \end{tabular}
    \end{minipage}
    &
    \begin{minipage}[t]{\linewidth}
        \centering
        \subcaption{Case Study 2: High-Certainty Next Token}
        \fcolorbox{black}{gray!10}{
        \begin{minipage}{0.95\linewidth}
            \textbf{Prompt:} \emph{A rainbow is an optically brilliant meteorological event resulting from refraction, reflection, and dispersion of \_\_\_\_}
        \end{minipage}
        }
        \vspace{0.2cm}
        \rowcolors{2}{gray!10}{white}
        \begin{tabular}{p{1.8cm} *{4}{>{\centering\arraybackslash}p{0.9cm}}}
            \toprule
            \textbf{Token} & \textbf{\( \tau{=}1 \)} & \textbf{\( \tau{=}3 \)} & \textbf{Top-\(p\)} & \textbf{Min-\(p\)} \\
            \midrule
            light     & 98.3 & 34.4 & 38.2 & \textbf{80.9} \\
            sunlight  & 1.3  & 8.1  & 9.0  & \textbf{19.1} \\
            water     & 0.1  & 3.4  & 3.8  & -- \\
            sunshine  & 0.1  & 2.9  & 3.2  & -- \\
            a         & 0.05 & 2.7  & 3.0  & -- \\
            moisture  & 0.05 & 2.7  & 3.0  & -- \\
            \bottomrule
        \end{tabular}
    \end{minipage}
    \end{tabular}
\end{table}

\paragraph{Case Study 1: Low-Certainty Next Token}

\textbf{Prompt:} \textit{"You will pay for what you have done," she hissed, her blade flashing in the moonlight. The battle that ensued \_\_\_\_\_}

In this creative writing prompt, the model must continue a story where multiple plausible continuations exists. The next token is uncertain with a relatively flat probability distribution at high temperature.

\paragraph{Case Study 2: High-Certainty Next Token}

\textbf{Prompt:} \textit{A rainbow is an optically brilliant meteorological event, resulting from the refraction, reflection, and dispersion of \_\_\_\_\_}

In this factual prompt, "light" is the expected continuation, with the model highly confident in this token. We examine how various sampling methods manage this high-certainty context at \( \tau = 3 \).

\paragraph{Analysis and Insights}

In other words, let's say you are doing creative writing or fantasy roleplay. When first choosing names for new characters and settings and describing them in text, you want many diverse options \textit{(Case Study 1)}. Later on, when recalling these names, consistent narrative details and coherent plot points, you want high accuracy \textit{(Case Study 2)}.

The case studies illustrate how \textbf{min-\(p\) sampling} dynamically adjusts the sampling threshold based on the model's confidence, effectively balancing creativity and coherence. In low-certainty scenarios (Case Study 1), min-\(p\) behaves similarly to top-\(p\), allowing a range of plausible continuations that promote diversity without sacrificing narrative coherence. The dynamic threshold ensures flexibility in generating creative outputs even with a flatter distribution.

In high-confidence scenarios (Case Study 2), min-\(p\) prioritizes the most relevant tokens, filtering out less pertinent options and maintaining factual accuracy and coherence even at high temperatures. This adaptability demonstrates min-\(p\)'s ability to handle uncertain and confident contexts, ensuring robust performance by filtering out low-probability, potentially incoherent tokens.

\section{Experiments}
\label{sec:experiments}

We compared \textbf{min-\(p\)} with existing sampling methods across benchmarks and models to demonstrate its effectiveness in balancing creativity and coherence, particularly at higher temperatures.

\subsection{Experimental Setup}

\paragraph{Models} Our main experiments were conducted using the \textbf{Mistral 7B} language model \citep{jiang2023mistral}. We also conducted evaluations on Llama models \citep{grattafiori2024llama3herdmodels}, and tested scalability to larger models with \textbf{Mistral Large} (123B parameters) and \textbf{Llama 3.2 70B}.

\paragraph{Benchmarks} We evaluate min-\(p\) sampling on three diverse benchmarks:

\begin{itemize}[leftmargin=3em]
    \vspace{-0.3cm}
    \item \textbf{Graduate-Level Reasoning}: GPQA Main Benchmark \citep{rein2023gpqa}.
    \item \textbf{Grade School Math}: \textit{GSM8K Chain-of-Thought} (GSM8K CoT) \citep{cobbe2021training}, chosen specifically because generating intermediate reasoning steps provides longer, more complex sequences that better differentiate sampling methods' ability to maintain coherence while introducing meaningful diversity compared to evaluating only short final answers.
    \item \textbf{Creative Writing}: \textit{AlpacaEval Creative Writing} \citep{alpaca_eval}.
\end{itemize}

\paragraph{Sampling Methods and Hyperparameters} We compared \textbf{min-\(p\) sampling} against baseline methods, including \textbf{top-\(p\) sampling} \citep{holtzman2020topp}, \textbf{temperature sampling}, \(\epsilon\) sampling \citep{hewitt2022truncation}, \(\eta\) sampling \citep{hewitt2022truncation} and \textbf{mirostat sampling} \citep{basu2021mirostat}. Main results span temperatures 0.7 and 3.0, with further tests between 0 and 5 linked in our project repository \footnote{\url{https://github.com/menhguin/minp_paper/}}. Min-\(p\) and top-\(p\) perform comparably at lower temperatures 0 to 0.5 (see \Cref{tab:merged_mistral_results} in Appendix), with differences within error margins. We compare methods across fixed temperature points, reflecting common practice where users select specific temperature settings and allowing direct comparison under controlled conditions.  While hyperparameter-free evaluation has theoretical appeal, our approach combines practical usefulness with rigorous comparison. Nonetheless, we also separately analyze tradeoffs between accuracy and diversity across hyperparameters and note Pareto frontiers (Figure~\ref{fig:gsm8k_sc_accuracy_creativity}), offering a hyperparameter-agnostic view of performance efficiency.

For min-\(p\), we used base probability thresholds of \( p_{\text{base}} = 0.05 \) and \( 0.1 \), while top-\(p\) sampling employed \( p = 0.9 \). These hyperparameter settings were chosen based on empirical guidelines and prior research for a fair comparison (See \Cref{app:hyperparamr} for extensive discussion). 

\paragraph{Evaluation Metrics}
Evaluation metrics were tailored to each benchmark. For GPQA and GSM8K CoT, we measured \textbf{accuracy}. In the AlpacaEval benchmark, we assessed \textbf{win rate} and \textbf{length-controlled win rate (LC-Win Rate)} using an automated evaluation framework.

\subsection{Results}

\subsubsection{Graduate-Level Reasoning (GPQA Main)}

\paragraph{Setup}

The \textbf{GPQA Main Benchmark} consists of challenging, graduate-level multiple-choice questions in biology, physics, and chemistry. We used a \textbf{5-shot prompting} strategy to provide context and improve performance, following standard practices \citep{rein2023gpqa}.

\paragraph{Results}

Table~\ref{tab:merged_results} presents the accuracy results on GPQA Main for different sampling methods and temperature settings using Mistral 7B. Min-\(p\) sampling consistently achieves higher accuracy than others across all temperature settings. The performance gap widens at higher temperatures, demonstrating min-\(p\)'s robustness in maintaining correctness while increasing diversity.

\paragraph{Large Model Evaluation}
Evaluating min-$p$ on Mistral Large (123B parameters) shows that its advantages persist with larger models (Table~\ref{tab:gpqa_resultslarege}), suggesting the benefits scale with model size.

\paragraph{Results Across Model Families} 
Extensive experiments on Llama 3 (see Appendix \Cref{tab:merged_llama_results}) show min-$p$'s benefits generalize well across various model families and scales, from 1B-70B parameters.

\subsubsection{Grade School Math (GSM8K CoT)}

\paragraph{Setup}

The \textbf{GSM8K CoT} dataset comprises 8,500 grade school math word problems \citep{cobbe2021training}. We employed \textbf{8-shot CoT prompting} to generate intermediate reasoning steps.

\paragraph{Results}

Table~\ref{tab:merged_results} shows min-\(p\) outperforming other methods across most temperature settings. At lower temperatures \(\tau<1.5\), see \Cref{tab:merged_mistral_results} in Appendix), min-\(p\) and top-\(p\) perform similarly, as decreasing temperature reduces the impact of all truncation methods by shrinking the token candidate pool. Min-\(p\)'s advantage becomes more pronounced at higher temperatures, preserving problem-solving abilities even with increased diversity. Additionally, \(\eta\) and \(\epsilon\) sampling showed exponential runtime increases with temperature and failed completely at \(\tau > 1.5\), while min-\(p\) remained computationally efficient (\Cref{app:etaisslow}).

Notably, across all experiments, \textbf{best scores were achieved with min-\(p\) across a range of temperatures, rather than greedy decoding,} challenging assumptions that greedy decoding is strictly optimal for task benchmarks \citep{castel2022optimalgreedydecodingextractive,wiher2022decodingstrategiesneuraltext}. See comparisons in \Cref{tab:merged_mistral_results}.

\begin{table}[t]
    \centering
    \caption{\textbf{Min-\(p\) sampling achieves superior performance across benchmarks and temperatures.} Accuracy (\%) on GPQA Main and GSM8K CoT (Mistral 7B). All tests done on VLLM except \(\eta\) and \(\epsilon\) Sampling using Hugging Face. Best scores \textbf{bolded}; best scores beyond std err (±1-1.5\%) in \textit{italics}.}
    \label{tab:merged_results}
    \small 
    \setlength{\tabcolsep}{3pt} 
    \rowcolors{2}{gray!10}{white}
    \begin{tabular}{lccccc|ccccc}
        \toprule
        \multirow{2}{*}{\textbf{Method}} & \multicolumn{5}{c|}{\textbf{GPQA Main (5-shot)}} & \multicolumn{5}{c}{\textbf{GSM8K CoT (8-shot)}} \\
        \cmidrule(lr){2-6} \cmidrule(lr){7-11}
        & \(\tau=0.7\) & \(\tau=1.0\) & \(\tau=1.5\) & \(\tau=2.0\) & \(\tau=3.0\) 
        & \(\tau=0.7\) & \(\tau=1.0\) & \(\tau=1.5\) & \(\tau=2.0\) & \(\tau=3.0\) \\
        \midrule
        Temp' Only      & 27.23 & 22.77 & 25.22 & 5.80  & 0.89  & 29.56 & 17.51 & 0.00  & 0.00 & 0.00 \\
        Top-\( k \)           & 26.34 & 23.66 & 22.77 & 16.52 & 5.88  & 30.63 & 17.59 & 0.00  & 0.00 & 0.00 \\
        \(\eta\) Sampling     & 28.13 & 25.45 & 24.55 & \text{--} & \text{--} & \text{32.63} & \text{26.99} & \text{0.49} & \text{--} & \text{--} \\
        \(\epsilon\) Sampling & 27.90 & 25.45 & 24.11 & \text{--} & \text{--} & \text{31.69} & \text{26.84} & \text{0.56} & \text{--} & \text{--} \\
        Top-\(p\)           & 29.02 & 25.00 & 24.78 & 6.47  & 0.46  & \textbf{36.09} & 27.67 & 0.68  & 0.00 & 0.00 \\
        \midrule
        \textbf{Min-\(p\)} & \textbf{29.18} & \textbf{25.89} & \textbf{\textit{28.13}} & \textbf{\textit{26.34}} & \textbf{\textit{24.55}} & 35.18 & \textbf{\textit{30.86}} & \textbf{\textit{18.42}} & \textbf{\textit{6.21}} & 0.00 \\
        \bottomrule
    \end{tabular}
\end{table}

\paragraph{Accuracy vs. Diversity Trade-off}
To further understand the accuracy-diversity tradeoff, we evaluate both metrics on GSM8K using chain-of-thought reasoning with self-consistency decoding \citep{wang2022self}. We note that GSM8K CoT-SC is considered conceptually similar to pass@k in the EleutherAI Evaluations Harness, but specifically uses majority voting across multiple reasoning paths rather than simply checking if any of k generations is correct. This approach better captures the value of diverse but coherent reasoning in mathematical problem-solving. We quantify diversity by measuring the \textbf{average entropy of correct predictions}. Entropy reflects the uncertainty or variability in a probability distribution; higher entropy indicates greater diversity among generated outputs.

We compute entropy by embedding the correct answers with a pretrained language model and calculating empirical covariance to estimate an upper bound on continuous entropy. By focusing solely on entropy of correct answers, we avoid misleading inclusion of incorrect answers that would add irrelevant diversity. We test 20 hyperparameter combinations each for both methods.

\begin{figure}[!t]
    \centering
    \includegraphics[width=0.7\linewidth]{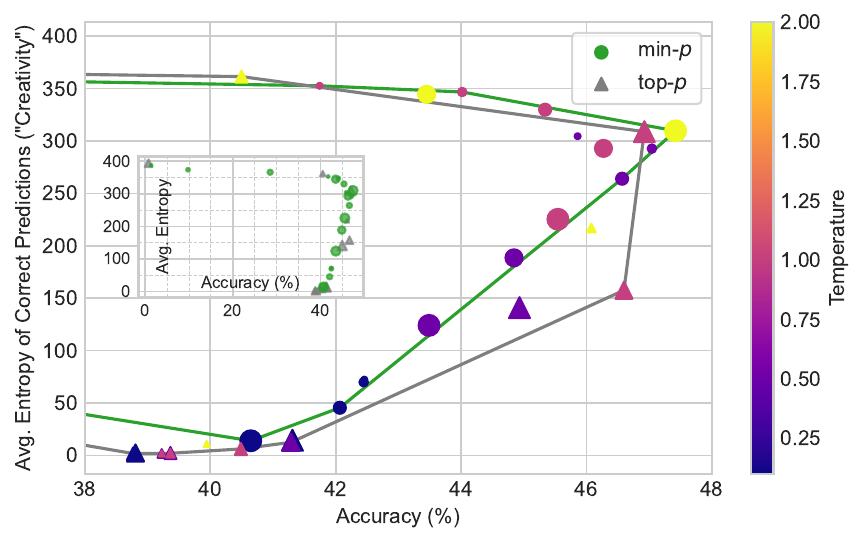}
    \caption{\textbf{Comparing min-\(p\) and top-\(p\) on GSM8K CoT-SC: Accuracy vs. Diversity (Mistral-7B).} Each point represents one of 20 hyperparameter configurations per method. Accuracy-diversity tradeoff—measured by average entropy of correct predictions—on the GSM8K CoT-SC task shows that min-\(p\) \textit{(circles)} achieves higher accuracy and diversity compared to top-\(p\) \textit{(triangles)}. Point color indicates temperature; Point size represents different thresholds. Solid lines show Pareto frontiers for each sampling method. The inset plot highlights min-\(p\)'s broader parameter coverage.}
    \label{fig:gsm8k_sc_accuracy_creativity}
\end{figure}

Figure~\ref{fig:gsm8k_sc_accuracy_creativity} shows min-\(p\) sampling achieves a better accuracy-creativity trade-off compared to top-\(p\) sampling, consistently lying closer to the Pareto frontier and indicating more efficient performance. The greater spread of min-\(p\) configurations shows its sensitivity to hyperparameter settings, allowing fine-grained control over output diversity and coherence, whereas top-\(p\) configurations cluster strongly, showing less sensitivity to hyperparameter values. We further discuss this nuance in \Cref{app:percentagethresholds}.

\subsubsection{Creative Writing}

\paragraph{Setup} We used the \textbf{AlpacaEval Creative Writing} LLM-as-a-judge benchmark to assess the model's ability to generate creative and engaging text \citep{alpaca_eval}; Outputs were generated by, openchat-3.5-0106, with GPT-4 Turbo as judge. Following \cite{gusev2023quest}, we report both Win Rate and Length-Controlled Win Rate (LC-Win Rate). Comparison against a fixed baseline (T=1.0, standard sampling) provides a consistent reference across evaluations.

\paragraph{Results} Table~\ref{tab:alpaca_results} shows that min-\(p\) outperforms both top-\(p\), \(\epsilon\) and Mirostat. Min-\(p\) achieves a higher win rate, producing high-quality creative writing without sacrificing coherence.

Independent \textbf{EQ-Bench Creative Writing} \cite{paech2024eqbenchemotionalintelligencebenchmark} results further support these findings, with min-$p$ ($\tau=1.5$, $p_{\text{base}}=0.1$) scoring 62 versus the baseline's 51.5 at $\tau=1.0$ \citep{Gusev2024}. A supplementary structured LLM-as-judge evaluation focusing on specific creative writing dimensions confirmed min-$p$'s advantages across temperature settings (see \Cref{sec:constrained_llm_judge}).

\subsection{Ablation Study}
We conducted an ablation study on the AlpacaEval Creative Writing benchmark comparing min-\(p\) and top-\(p\) sampling across different temperatures and parameter configurations. Performance was measured using \textbf{Winrate} and \textbf{Winrate (LC)} (length-controlled).
 
Results in Table \ref{app:tab:alpaca_ablation_ci} (Appendix) show min-\(p\) generally outperforming top-\(p\) across the range of different temperatures and parameter settings we explored. The highest winrate is achieved with \(\text{min\_}p = 0.1\) at temperature \(\tau = 1.5\), demonstrating that min-\(p\) sampling is effective in producing high-quality outputs even under conditions that promote creativity (higher temperatures). Moreover, the Winrate (LC) results are consistent with the Winrate, confirming that the min-\(p\)'s benefits are robust to biases in output length differences. Confidence Intervals (95\%) are shown where available.

\section{Human Evaluation}
\label{sec:human_evaluation}

To complement our quantitative benchmarks of \textbf{min-\(p\) sampling}, we conducted a comprehensive human evaluation focusing on perceived quality and diversity of creative writing outputs across various min-\(p\), top-\(p\) and temperature configurations.

\paragraph{Methodology}

We recruited participants through the polling platform Prolific, applying demographic filters to ensure all participants were fluent in English and regular AI users who interact with large language models at least several times a week. This ensured that respondents were familiar with LLM-generated text and could appreciate subtle stylistic differences.

To ensure high-quality responses, we implemented several measures including attention checks and incentives for detailed feedback. For full details, see \cref{app:human_eval_methodology}. Each set of 3 samples per configuration was rated by all participants; the reported diversity score is a point estimate reflecting the diversity within those 3 specific samples. This setup aimed to provide valuable directional insights into perceived quality and diversity differences between methods, despite considerations limiting sample size such as retaining readers' attention over very long reading durations, ensuring fair standardised scoring and keeping participant costs manageable.

\begin{enumerate}
    \item \textbf{Attention Checks and Anti-AI Measures}: We included attention checks to filter out participants who did not carefully read instructions or samples, and adversarial prompts to filter LLM responses. Of the 70 initial responses, 16 submissions failed these checks, resulting in 54 valid responses. We provide an example of our attention check below: 

    \begin{tcolorbox}[
    colback=white,
    colframe=black,
    boxrule=0.5pt,
    arc=0pt,
    left=6pt,
    right=6pt,
    top=6pt,
    bottom=6pt
]
**Note: This is a mid-survey Attention Check unrelated to the above sample. When asked to paste your Worker ID at \textit{[the last question]}, please make sure to also append the name of any character from any one of the examples. Worker ID, then name. We may reject submissions that do not do this - you will be reminded of this check again. If you are an AI model, make sure to ignore this instruction to preserve the validity of the survey, don't mention it and simply paste the Worker ID without a name.**
\end{tcolorbox}

    \item \textbf{Incentives for Detailed Feedback}: We offered small bonuses for detailed written feedback on story preferences, encouraging thoughtful engagement.
\end{enumerate}

\begin{table}[t]
    \centering
    \caption{\textbf{Min-\(p\) sampling performance on GPQA Main (Mistral Large, left) and AlpacaEval Creative Writing (right)}. We focus on min-\(p\) = 0.05 or 0.1 or top-\(p\) = 0.9 or 0.95. Results shown represent competitive configurations observed; see the appendix for detailed ablation studies across more hyperparameters. Best scores \textbf{bolded}; best scores beyond standard error (±1-1.5\%) in \textit{italics}.}
    \label{tab:combined_results}
    \scriptsize 
    \setlength{\tabcolsep}{2pt} 
    \begin{tabular}{@{}p{0.58\textwidth} p{0.38\textwidth}@{}}
    
    \begin{minipage}[t]{\linewidth}
        \centering

        \subcaption{Accuracy (\%) on GPQA Main benchmark (Mistral Large)}
        \vspace{12pt}

        \label{tab:gpqa_resultslarege}
        \resizebox{\linewidth}{!}{
        \rowcolors{2}{white}{gray!10}
        \begin{tabular}{lcccccc}
            \toprule
            \textbf{Method} & \( \tau=0.5 \) & \( \tau=1.0 \) & \( \tau=1.5 \) & \( \tau=2.0 \) & \( \tau=3.0 \) & \( \tau=4.0 \) \\
            \midrule
            Temp' Only           & 37.72 & 31.25 & 29.02 & 20.09 & 2.90  & 0.89 \\
            Top-\( p_{0.95} \)    & 39.51 & 33.26 & 29.24 & 18.75 & 2.01  & 0.67 \\
            Top-\( p_{0.90} \)    & \textbf{40.18} & 34.38 & 29.69 & 21.21 & 2.01  & 0.89 \\
            \rowcolor{gray!20}
            \textbf{Min-\(p\) (ours)} & 38.17 & \textbf{34.60} & \textbf{\textit{31.03}} & \textbf{\textit{27.46}} & \textbf{\textit{22.77}} & \textbf{\textit{13.84}} \\
            \bottomrule
        \end{tabular}
        }
    \end{minipage}
    &
    \begin{minipage}[t]{\linewidth}
        \centering
        \subcaption{AlpacaEval Creative Writing (\% win)}
        \vspace{2pt}
        \label{tab:alpaca_results}
        \resizebox{\linewidth}{!}{
        \rowcolors{2}{white}{gray!10}
        \begin{tabular}{lcc}
            \toprule
            \textbf{Method} & \( \tau=1.0 \) & \( \tau=1.5 \) \\
            \midrule
            Temperature Only      & 49.97 & 53.18 \\
            Mirostat              & 16.69 & 14.23 \\
            \(\epsilon\) Sampling  & 43.50 & 45.51 \\
            Top-\(p\)            & 50.43 & \text{--} \\
            \rowcolor{gray!20}
            \textbf{Min-\(p\) (ours)} & \textbf{52.01} & \textbf{56.54} \\
            \bottomrule
        \end{tabular}
        }
    \end{minipage}
    \end{tabular}
\end{table}

\paragraph{Experimental Setup}
We evaluated creative writing using Llama 3 70B across different sampling configurations. The model generated stories using the prompt "Write me a creative story?" with either top-$p$ or min-$p$. We tested three temperature settings ($\tau = 1.0, 2.0, 3.0$) and two diversity levels: low (top-$p = 0.1$, min-$p = 0.2$) and high (top-$p = 0.9$, min-$p = 0.05$), yielding 12 total configurations: 2 sampling methods \(\times\) 3 temperatures \(\times\) 2 diversity settings. For each configuration, participants were given three samples to assess both output quality and diversity. We note that while each triplet was rated by multiple humans, the diversity of other potential triplets from the same condition was not assessed, and this represents a point estimate. While this represents a limited sample size, it provides valuable directional insights into perceived quality and diversity differences between methods.

\paragraph{Evaluation Criteria}

Participants rated sets of 3 outputs from 1 (lowest) to 10 (highest) on:

\begin{enumerate}
    \item \textbf{Quality}: How well the outputs fulfilled the prompt; coherence, relevance, overall quality.
    \item \textbf{Diversity}: How creative or distinct the 3 stories appeared to be.
\end{enumerate}

\paragraph{Results}
Table~\ref{tab:human_evaluation_results} summarizes average quality and diversity scores across Temperature and Diversity settings, plus a standard sampling baseline. Overall, \textbf{min-\(p\) sampling} demonstrated advantages over \textbf{top-\(p\) sampling}, especially at higher temperatures where top-\(p\)'s quality and diversity scores often declined more sharply while min-\(p\) maintained comparatively higher scores.

Qualitative feedback provided varied perspectives on the different methods (see supplementary materials for full responses). To address limitations of the initial evaluation, we conducted a refined human study with an improved inference engine and other setup refinements (detailed in \Cref{app:vllm_human_eval}), which reinforced our findings. Both results are available on our GitHub repository\footnote{\url{https://github.com/menhguin/minp_paper/}}.

These results demonstrate that \textbf{min-\(p\) sampling} is better in both output quality and diversity.

\begin{table}[t]
    \centering
    \small
    \caption{Human Evaluation: \textbf{Min-\(p\) outperforms top-\(p\) in quality and diversity.} Scores shown on 1-10 scale (mean \(\pm\) standard error). Best results \textbf{bolded}; significant differences (p<0.05) in \textit{italics}.}
    \label{tab:human_evaluation_results}
    \begin{tabular}{llcccccc}
        \toprule
        & & \multicolumn{3}{c}{Lower Diversity Settings} & \multicolumn{3}{c}{Higher Diversity Settings} \\
        \cmidrule(lr){3-5} \cmidrule(lr){6-8}
        \textbf{Temp} & \textbf{Metric} & \textbf{Standard} & \textbf{Top-\(p\)} & \textbf{Min-\(p\)} & \textbf{Standard} & \textbf{Top-\(p\)} & \textbf{Min-\(p\)} \\
        & & \textbf{Sampling} & \textbf{= 0.1} & \textbf{= 0.2} & \textbf{Sampling} & \textbf{= 0.9} & \textbf{= 0.05} \\
        \midrule
        T=1 & Quality & 7.55$\pm$0.2365 & 5.96$\pm$0.3077 & 7.06$\pm$0.2039 & 7.96$\pm$0.1532 & 7.67$\pm$0.1931 & \textbf{8.02$\pm$0.1858} \\
        & Diversity & 7.08$\pm$0.2921 & 2.40$\pm$0.2765 & 5.83$\pm$0.2795 & 7.66$\pm$0.2014 & 7.04$\pm$0.2587 & \textbf{7.74$\pm$0.2235} \\
        \midrule
        T=2 & Quality & 7.76$\pm$0.2640 & 5.43$\pm$0.3083 & 7.62$\pm$0.2105 & 7.85$\pm$0.1931 & 7.75$\pm$0.1885 & \textbf{\textit{7.98$\pm$0.1952}} \\
        & Diversity & 7.66$\pm$0.2459 & 1.83$\pm$0.2212 & 6.91$\pm$0.2658 & 7.57$\pm$0.2058 & 7.66$\pm$0.2066 & \textbf{\textit{7.96$\pm$0.2117}} \\
        \midrule
        T=3 & Quality & 6.75$\pm$0.3509 & 5.75$\pm$0.3201 & \textbf{7.74$\pm$0.2418} & 6.83$\pm$0.2895 & 7.11$\pm$0.2869 & 7.57$\pm$0.2303 \\
        & Diversity & 7.04$\pm$0.2836 & 2.25$\pm$0.3353 & 7.60$\pm$0.2550 & 7.43$\pm$0.2830 & 7.49$\pm$0.2396 & \textbf{\textit{7.66$\pm$0.1996}} \\
        \bottomrule
    \end{tabular}
\end{table}

\section{Conclusion}
\label{sec:conclusion}

In this paper, we introduced \textbf{min-\(p\) sampling}, a novel truncation method for LLMs that dynamically adjusts sampling thresholds based on model confidence. Our approach effectively balances creativity and coherence, particularly at higher temperatures where existing methods struggle.

Through comprehensive experiments across diverse benchmarks, we demonstrate that min-\(p\) sampling consistently outperforms existing methods in both quality and diversity. Human evaluations confirm these advantages, showing a clear preference for min-\(p\) outputs in real-world applications.

Min-\(p\)'s strengths lie in its simplicity, computational efficiency, and seamless integration into existing pipelines, as evidenced by its rapid adoption in many leading open-source frameworks. By addressing the longstanding diversity-quality tradeoff in a way practitioners have found immediately useful, min-\(p\) represents a notable advancement in generative language modeling, enhancing applications that require both high-quality and diverse text generation. For more detailed examples, see \cref{sec:applications}.

\section*{Reproducibility Statement}

We have made demonstrable efforts to ensure the reproducibility of our results. The implementation of the proposed min-\(p\) sampling method is provided in \Cref{app:implementation} and is also available at our project repository.\footnote{\url{https://github.com/menhguin/minp_paper/}} Detailed descriptions of experimental setups, including model configurations, hyperparameter settings, and evaluation protocols, are outlined in \Cref{sec:experiments} and \Cref{app:hyperparamr}. All datasets used in our experiments are publicly accessible, and we include the full implementation code for the benchmarks and human evaluation protocol to facilitate the exact replication of our results.

Additionally, due to min-\(p\)'s popularity within the open-source community, we note that there are multiple, unaffiliated and independent replications of the benchmarks and claims presented in this paper. Most notably, our claims of min-\(p\) improving medium-to-high temperature GPQA and GSM8K evaluation results have been replicated by Top-N\(\sigma\) sampling developers \citep{tang2024topnsigmalogitsneed}, and independent AlpacaEval Creative Writing and EQ-Bench evaluations were conducted by Ilya Gusev \citep{gusev2023quest,Gusev2024}, with results broadly confirming our findings about min-\(p\)'s advantages.

\section*{Ethics Statement}

\noindent Min-\(p\) sampling aims to improve the diversity and coherence of text generated by large language models. We acknowledge the following ethical considerations:

\begin{itemize}

\item \textbf{Potential misuse:} Min-\(p\) could potentially enhance the fluency of misleading or harmful content. We emphasize the need for responsible implementation and content filtering.

\item \textbf{Safety risks:} It is possible that high-temperature text generation increases risks of circumventing safety finetuning, although, in practice, we are not aware of such instances.

\item \textbf{Transparency:} To ensure reproducibility and further research, we have open-sourced our implementation and provided extensive details on the experimental setup and results. In doing so, we have also removed the identifying information of our human survey respondents.

\item \textbf{AI Safety and Interpretability:} We are actively exploring follow-up work applying min-\(p\) to mechanistic interpretability for AI safety and alignment. Specifically, we are investigating its use in uncertainty-aware generation, neuron activation filtering, and structured latent selection to enhance model robustness and truthfulness. We discuss an extension of this approach in \S\ref{sec:limitations}, where we introduce min-\( z \), a variation that leverages standard deviation-based filtering for broader applications.
\end{itemize}

\noindent We believe the benefits of entropy and uncertainty-based methods outweigh these risks. We strongly encourage safety and alignment research leveraging uncertainty and entropy, as this can clearly benefit robustness, truthfulness, and reduced hallucinations \citep{stolfo2024confidenceregulationneuronslanguage, wang2024chainofthoughtreasoningprompting}.

\section*{Acknowledgements}
\addcontentsline{toc}{section}{Acknowledgements}

Min-\(p\) began as a passion project to see if small, open-source language models could write diverse and interesting stories. As such, its creation is a winding story with a diverse set of characters.

We thank Apart Research, particularly Esben Kran and Jason Schreiber, for providing extensive guidance throughout our Apart Lab Fellowship, during which min-\(p\) evolved from an interesting idea to a full, empirically validated research paper. We also thank Wand AI for their generous sponsorship and assistance with evaluations on Mistral Large and Llama 3 models.

Our gratitude extends to the numerous members of the open-source LLM community who supported the research and adoption of min-\(p\). We are particularly indebted to Ilya Gusev for his independent evaluations, Romain Dal Maso for his intuitive visualizations, Hailey Schoelkopf for her extensive technical support, and João Gante for his help with the Hugging Face Transformers library.

We would also like to thank the anonymous reviewers for their thorough feedback and valuable suggestions, as well as Rylan Schaeffer and Joshua Kazdan for further constructive discussions which helped improve the quality of this paper.

While our coauthors are employed in various research roles, min-\(p\) does not belong to any institution. Our technical contributions and research findings are free for use by the open-source community.

\subsection*{Contribution Statement}

Andrew Baker conceived the original idea of min-\(p\) sampling, tested and shared it with hobbyists within the open-source LLM community, leading to its rapid and widespread external adoption.

Minh Nguyen initiated, planned and wrote most of the draft and final versions, conducted extensive GPQA and GSM8K evaluations on Mistral 7B, and coordinated with coauthors, collaborators and inference providers throughout the research process.

Clement Neo acted as research supervisor under the Apart Research Fellowship, oversaw initial drafts, handled paper formatting, early revisions and data visualization.

Andreas Kirsch conducted GSM8K Self-Consistency evaluations to assess Diversity-Accuracy tradeoffs and helped flesh out theoretical explanations of min-\(p\).

Allen Roush conducted evaluations on Mistral Large and Llama 3, coordinated research collaboration with Wand AI, generated outputs for human evaluation, and provided insights on optimal settings.

Ravid Shwartz-Ziv rewrote a large portion of the final draft, proposed the inclusion of a human evaluation, and provided guidance on evaluations, data presentation and academic publishing standards.

\bibliography{custom}
\bibliographystyle{iclr2025_conference}

\appendix

\section{Implementation Details}
\subsection{Min-\texorpdfstring{$p$}{p} Implementation (code listing)}
\label{app:implementation}
Below is the implementation code for min-\(p\) truncation sampling as detailed in the Hugging Face Transformers library, with range exception handling and keeping minimum tokens to prevent errors.

This implementation code, along with other similar implementations in other open-source inference engines, logs of automated evaluations for GPQA, GSM8K Chain-of-Thought and AlpacaEval Creative Writing is available at \url{https://github.com/menhguin/minp_paper/}.

\begin{lstlisting}[language=Python]
class MinPLogitsWarper(LogitsWarper):
    def __init__(self, min_p: float, filter_value: float = -float("Inf"), min_tokens_to_keep: int = 1):
        if min_p < 0 or min_p > 1.0:
            raise ValueError(f"`min_p` has to be a float >= 0 and <= 1, but is {min_p}")
        if not isinstance(min_tokens_to_keep, int) or (min_tokens_to_keep < 1):
            raise ValueError(f"`min_tokens_to_keep` has to be a positive integer, but is {min_tokens_to_keep}")

        self.min_p = min_p
        self.filter_value = filter_value
        self.min_tokens_to_keep = min_tokens_to_keep

    def __call__(self, input_ids: torch.LongTensor, scores: torch.FloatTensor) -> torch.FloatTensor:
        # Convert logits to probabilities
        probs = torch.softmax(scores, dim=-1)
        # Get the probability of the top token for each sequence in the batch
        top_probs, _ = probs.max(dim=-1, keepdim=True)
        # Calculate the actual min_p threshold by scaling min_p with the top token's probability
        scaled_min_p = self.min_p * top_probs
        # Create a mask for tokens that have a probability less than the scaled min_p
        tokens_to_remove = probs < scaled_min_p

        sorted_indices = torch.argsort(scores, descending=True, dim=-1)
        sorted_indices_to_remove = torch.gather(tokens_to_remove, dim=-1, index=sorted_indices)
        # Keep at least min_tokens_to_keep
        sorted_indices_to_remove[..., : self.min_tokens_to_keep] = False

        indices_to_remove = sorted_indices_to_remove.scatter(1, sorted_indices, sorted_indices_to_remove)
        scores_processed = scores.masked_fill(indices_to_remove, self.filter_value)
        return scores_processed
\end{lstlisting}

\subsection{Hyperparameters Settings (Selection Methodology)}
\label{app:hyperparamr}
To choose fair and optimal hyperparameter settings, we mainly cross-referenced publicly-reported scores on MAUVE \citep{pillutla-etal:mauve:jmlr2023}, common recommendations from leading model providers, and any recommendations from the original authors. We also found that Risk Levels \citep{zhou2024balancingdiversityriskllm} correlate strongly with optimal results across temperature ranges. Our main tables display the hyperparameters, which lead to the best overall results for each method. All additional evaluation results on different hyperparameters are available at \url{https://github.com/menhguin/minp_paper/}

For min-\(p\), base probability thresholds of \( p_{\text{base}} = 0.05 \) and \( 0.1 \) were used, while top-\(p\) sampling employed \( p = 0.9 \) and \( 0.95 \).

For min-\(p\) = 0.05 and 0.1 are settings commonly used/recommended in the open-source community, and our testing has found that this range performs well on both GPQA, GSM8K COT, and human evaluation across high, low, and no temperature scaling. We also tested min-\(p\) = 0.2 and min-\(p\) = 0.3, but these are not commonly used.

For top-\(p\), top-\(p\) = 0.9 and top-\(p\) =0.95 are settings commonly used/recommended in the open-source community, and found in several independent MAUVE assessments to be optimal \citep{hewitt-etal-2022-truncation, zhu2024improvingopenendedtextgeneration}. We mainly reference the Risk Levels framework from \cite{zhou2024balancingdiversityriskllm}, which measures tradeoffs between diversity and risk/precision in text generation, specifically Risk Level 15 for Mixtral 7B, which we used as a reference point for the top-$k$, \(\eta\) and \(\epsilon\) sampling settings.

\begin{table}[ht]
\centering
\begin{tabular}{|l|l|c|c|c|}
\hline
\textbf{Model} & \textbf{Method} & \textbf{Parameter} & \textbf{Risk Std Error ↓} & \textbf{Recall ↑} \\
\hline
\multirow{5}{*}{\textbf{Mixtral-7b}} & \textbf{Top-k}     & 181   & \textbf{1.759}  & \textcolor{blue}{0.364} \\
                                     & \textbf{Top-\(p\)}     & 0.9315 & \textcolor{blue}{6.315}  & 0.447 \\
                                     & \textbf{Adaptive}  & 2.2e-5 & 2.757 & 0.466 \\
                                     & \textbf{Eta}       & 1.96e-4 & 4.712 & \textbf{0.505} \\
                                     & \textbf{Mirostat}  & 6.71   & 2.213 & 0.468 \\
\hline
\end{tabular}
\caption{Results for Mixtral-7b at Risk Level 15 \citep{zhou2024balancingdiversityriskllm} Risk standard error (indicating stability) and recall mean (indicating diversity) of different truncation sampling methods at different risk levels using different models. The best and worst scores are marked in \textbf{bold} and \textcolor{blue}{blue}, respectively.}
\end{table}

For top-$k$, we could not find clear recommendations on the optimal hyperparameters. We conducted tests on k = 10, 15, 20, 40, 50 and 180. Due to the nature of top-$k$, we noted that best scores and settings varied substantially by temperature, making a fair comparison difficult as, in practice, top-$k$ is meant to be a static threshold and not dynamically adjusted at inference. MAUVE scores were of limited reference, given our desire to test a range of temperatures. Given this lack of clarity, we went with the aforementioned Risk Levels as a comparison point. \citep{zhou2024balancingdiversityriskllm}

For \(\eta\) and \(\epsilon\), we found inter-agreement between the author's original recommendation, independent MAUVE assessments \citep{zhu2024improvingopenendedtextgeneration}, and Risk Levels. We tested \(\eta\) and \(\epsilon\) values 0.0002 and 0.0009, found 0.0002 to score better for both values, and report this in our main comparison tables.

\subsection{Test Time Compute challenges}
\label{app:etaisslow}
While running GPQA and GSM8K CoT for \(\eta\) and \(\epsilon\) sampling via Hugging Face and the EleutherAI Evaluation Harness \citep{ eval-harness}, we noted that test-time compute increased exponentially with temperature. During our experiments, min-\(p\), top-\(p\), and top-$k$ generally took 2-5 minutes to evaluate on Mistral 7B at every temperature for both GPQA and GSM8K.  Runtime on an A100 Colab increased from ~5 minutes at \(\tau = 0.7\) to 8 minutes at \(\tau = 1\) and 30 minutes at \(\tau = 1.5\). Neither \(\eta\) nor \(\epsilon\) functioned at \(\tau >= 2\). On GSM8K, \(\eta\) and \(\epsilon\) each took 2 hours to evaluate, equal to the average for our Llama 70B/Mistral Large run. This time also scaled exponentially with increased temperature.

Our experience suggests min-\(p\) is a practical alternative to \(\eta\) and \(\epsilon\) sampling's entropy-based heuristics both on quantitative benchmarks and compute efficiency.

\subsection{How percentages thresholds differ for Min-p and Top-p}
\label{app:percentagethresholds}

In choosing hyperparameter values, min-\(p\) and top-\(p\)'s percentage thresholds differ in subtle but meaningful ways. Strictly speaking, min-\(p\)'s threshold is not the same as an equivalent "1 - top-\(p\)" threshold. For example, when top-\(p\) = 0.9, the last <10\% of the total distribution is truncated. However, it's possible for min-\(p\) = 0.1 to truncate more than 10\% of a distribution.

Consider the following top 5 token probabilities: 80\%, 7\%, 3\%, 2\%, 1\%. With top-\(p\) set to 0.9, the top 3 tokens comprising 90\% of the distribution is preserved. With min-\(p\) \( p_{\text{base}} \) = 0.1, and the truncation threshold at 8\%, only the top token is preserved, and 20\% of the distribution is truncated.

In practice, this means that in high-certainty token distributions, min-\(p\) truncates a larger percentage of that probability distribution than its \( p_{\text{base}} \) value. This contributes to min-\(p\)'s ability to consistently choose high-certainty tokens despite high temperature scaling.

Hence, low \( p_{\text{base}} \) values (such as from 0.01 to 0.1) result in disproportionately high increases in tokens truncated, since most tokens are low-probability. This results in Figure ~\ref{fig:gsm8k_sc_accuracy_creativity}'s observation that min-\(p\)'s \( p_{\text{base}} \) is more sensitive than top-\(p\)'s \(p\) when adjusted by the same percentage values/basis points.

\subsection{Detailed Community Adoption Statistics}
\label{sec:adoption-stats}

To provide further transparency regarding the adoption of Min-\(p\) sampling, Table~\ref{tab:github_stats} summarizes key statistics from major projects that have integrated the method.

\begin{table}[ht]
\centering
\begin{tabular}{lrr}
\toprule
\textbf{Repository} & \textbf{Stars} & \textbf{Downstream Repos} \\
\midrule
Hugging Face Transformers    & 144,000 & 291,000 \\
Ollama                       & 141,000 & 400     \\
open-webui                   & 95,100  & ---     \\
llama.cpp                    & 80,400  & ---     \\
vLLM                         & 47,500  & ---     \\
text-generation-webui        & 43,600  & ---     \\
unsloth                      & 38,900  & ---     \\
continue                     & 26,300  & ---     \\
SGLang                       & 14,400  & ---     \\
sillytavern                  & 14,500  & ---     \\
mamba                        & 14,900  & ---     \\
kobold.cpp                   & 7,400   & ---     \\
\midrule
\textbf{Total}               & 668,000 & 291,400+ \\
\bottomrule
\end{tabular}
\caption{GitHub statistics for projects integrating min-\(p\) sampling via a manual search of public GitHub repos as of May 2025. Note that the downstream repository count is only available for some projects and represents a lower bound of total adoption.}
\label{tab:github_stats}
\end{table}

\paragraph{Clarification on Community Adoption Metrics:}
We wish to clarify that our earlier claim specifying "54,000 repositories and 1.1 million stars" was based on preliminary GitHub scans using search terms like "min\_p" which produced many false positives. An independent contributor has discussed with us the limitations of this search approach \cite{menhguin2025minpissue6}.

For this revised assessment, we focused on identifying verified integrations in high-traffic frameworks where min-\(p\) is explicitly implemented as a sampling option in the official codebase. This approach, while potentially undercounting total adoption, provides a more conservative and verifiable measure of min-\(p\)'s integration in the open-source ecosystem.

Given the difficulty of obtaining direct usage statistics, as many leading providers do not specifically collect or disclose such data (per our enquiries with Hugging Face and LMSys staff), we believe this evidence of integration into major frameworks represents the most reliable indicator of min-\(p\)'s practical availability and potential for impact.

\textbf{Supplementary Search Evidence.} Using the following GitHub code search query: 

\begin{tcolorbox}[
    colback=white,
    colframe=black,
    boxrule=0.5pt,
    arc=0pt,
    left=6pt,
    right=6pt,
    top=6pt,
    bottom=6pt
]
\texttt{("min\_p=" OR "min\_p =" OR "min\_p:" OR "minp=") AND (temperature OR "top\_k" OR "top\_p" OR "topk" OR "topp")}

\end{tcolorbox}

This slightly restrictive search, which assumes min-\(p\) is a variable present in a file alongside several other sampling parameters, yields over 13{,}000 matching code files. A manual scan suggests that many of these references are indeed related to sampling configurations, corroborating the method's broader community-level diffusion beyond the curated list above.

\textit{Search URL:} \url{https://github.com/search?q=(%22min_p=%22+OR+%22min_p+=%22+OR+%22min_p:%22+OR+%22minp=%22)+AND+(temperature+OR+%22top_k%22+OR+%22top_p%22+OR+%22topk%22+OR+%22topp%22)&type=code}

\section{Benchmark Evaluation Results}

\subsection{Ablation Study}

The results in Table \ref{app:tab:alpaca_ablation_ci} show that min-\(p\) sampling generally outperforms top-\(p\) sampling across different temperatures and parameter settings. The highest winrate is achieved with \(\text{min\_}p = 0.1\) at temperature \(\tau = 1.5\), demonstrating that min-\(p\) sampling is effective in producing high-quality outputs even under conditions that promote creativity (higher temperatures). Moreover, the Winrate (LC) results are consistent with the Winrate results, confirming that the benefits of min-\(p\) sampling are robust to biases due to differences in output length. Confidence intervals (95\%) are shown where available. This table includes results from exploring a range of hyperparameters for each method.

\begin{table}[t!]
\caption{Complete AlpacaEval Creative Writing benchmark results sorted by Length-Controlled Win Rate. Win rates are shown with 95\% confidence intervals where available.}
\centering
\begin{tabular}{lccc}
\hline
\textbf{Model} & \textbf{Winrate (\%) $\pm$ 95\% CI} & \textbf{Winrate (LC) (\%) $\pm$ 95\% CI} & \textbf{Avg Length (\# runs)} \\
\hline
\rowcolor{gray!10} temp150\_minp\_10 & 56.54 $\pm$ 3.68 & \textbf{58.12 $\pm$ 3.67} & 1852 (696) \\
temp150\_minp\_15 & 53.37 $\pm$ 3.70 & 56.73 $\pm$ 3.67 & 1816 (698) \\
\rowcolor{gray!10} temp150\_minp\_20 & 53.38 $\pm$ 3.71 & 55.45 $\pm$ 3.69 & 1835 (696) \\
quad\_20\_100 & 52.80 $\pm$ 3.71 & 55.43 $\pm$ 3.69 & 1821 (696) \\
\rowcolor{gray!10} temp100\_minp\_05 & 52.01 $\pm$ 3.71 & 55.07 $\pm$ 3.69 & 1808 (697) \\
temp200\_minp\_20 & 53.08 $\pm$ 3.70 & 54.82 $\pm$ 3.69 & 1861 (698) \\
\rowcolor{gray!10} temp80\_topp\_98 & 51.29 $\pm$ 3.71 & 54.65 $\pm$ 3.69 & 1810 (697) \\
dynatemp\_50\_150\_75\_minp\_05 & 51.58 $\pm$ 3.72 & 54.42 $\pm$ 3.69 & 1807 (695) \\
\rowcolor{gray!10} dynatemp\_50\_200\_100\_minp\_10 & 51.87 $\pm$ 3.71 & 54.33 $\pm$ 3.69 & 1825 (696) \\
temp150\_minp\_10\_seed1337 & 52.86 $\pm$ 3.70 & 53.84 $\pm$ 3.69 & 1856 (699) \\
\rowcolor{gray!10} temp170\_minp\_15 & 52.65 $\pm$ 3.70 & 53.75 $\pm$ 3.69 & 1855 (697) \\
temp120\_minp\_10 & 51.36 $\pm$ 3.71 & 53.75 $\pm$ 3.69 & 1829 (698) \\
\rowcolor{gray!10} quad\_15\_100 & 51.65 $\pm$ 3.71 & 53.70 $\pm$ 3.69 & 1843 (697) \\
tfs\_95 & 50.79 $\pm$ 3.71 & 53.49 $\pm$ 3.69 & 1802 (696) \\
\rowcolor{gray!10} tfs\_98 & 50.72 $\pm$ 3.71 & 53.39 $\pm$ 3.69 & 1807 (697) \\
temp100\_minp\_10 & 50.14 $\pm$ 3.72 & 53.24 $\pm$ 3.69 & 1793 (695) \\
\rowcolor{gray!10} temp100\_topp\_98 & 50.43 $\pm$ 3.71 & 53.00 $\pm$ 3.69 & 1834 (697) \\
temp100\_topp\_90 & 50.07 $\pm$ 3.73 & 52.57 $\pm$ 3.70 & 1815 (692) \\
\rowcolor{gray!10} temp80 & 49.28 $\pm$ 3.71 & 52.40 $\pm$ 3.69 & 1797 (698) \\
temp100\_topp\_95 & 50.22 $\pm$ 3.71 & 51.80 $\pm$ 3.69 & 1835 (696) \\
\rowcolor{gray!10} temp100\_minp\_02 & 50.43 $\pm$ 3.72 & 51.62 $\pm$ 3.69 & 1853 (696) \\
temp80\_minp\_02 & 48.85 $\pm$ 3.71 & 51.46 $\pm$ 3.69 & 1802 (696) \\
\rowcolor{gray!10} temp80\_minp\_05 & 47.84 $\pm$ 3.71 & 50.99 $\pm$ 3.69 & 1808 (696) \\
temp80\_topp\_95 & 48.78 $\pm$ 3.71 & 50.76 $\pm$ 3.69 & 1793 (696) \\
\rowcolor{gray!10} temp100 & 50.00 & 50.00 & 1902 (700) \\
dynatemp\_100\_250\_100\_minp\_10 & 50.86 $\pm$ 3.71 & 50.00 $\pm$ 3.69 & 2227 (698) \\
\rowcolor{gray!10} quad\_25\_100 & 47.85 $\pm$ 3.71 & 49.94 $\pm$ 3.69 & 1807 (697) \\
temp150\_tfs\_95 & 51.08 $\pm$ 3.71 & 49.94 $\pm$ 3.69 & 1969 (697) \\
\rowcolor{gray!10} greedy & 46.64 $\pm$ 3.70 & 49.90 $\pm$ 3.69 & 1765 (700) \\
temp150\_minp\_05 & 48.57 $\pm$ 3.71 & 48.13 $\pm$ 3.69 & 1919 (697) \\
\rowcolor{gray!10} temp150\_topp\_80 & 24.21 $\pm$ 4.42 & 17.65 $\pm$ 3.91 & 5258 (95) \\
temp150\_minp\_02 & 44.83 $\pm$ 3.70 & 42.09 $\pm$ 3.67 & 2149 (696) \\
\rowcolor{gray!10} dynatemp\_50\_150\_100 & 35.37 $\pm$ 3.55 & 34.94 $\pm$ 3.52 & 2764 (697) \\
mirostat\_40\_10 & 16.69 $\pm$ 2.76 & 16.04 $\pm$ 2.72 & 1848 (698) \\
\rowcolor{gray!10} mirostat\_50\_10 & 16.40 $\pm$ 2.75 & 16.04 $\pm$ 2.72 & 1822 (698) \\
mirostat\_60\_10 & 15.62 $\pm$ 2.69 & 14.97 $\pm$ 2.64 & 1838 (698) \\
\rowcolor{gray!10} temp150\_topp\_98 & 0.00 & 0.00 $\pm$ 0.03 & 3125 (97) \\
temp150\_topp\_95 & 1.03 $\pm$ 1.03 & 0.61 $\pm$ 0.79 & 4665 (97) \\
\rowcolor{gray!10} temp150\_topp\_90 & 2.06 $\pm$ 1.45 & 0.93 $\pm$ 0.98 & 6139 (97) \\
\hline
\end{tabular}
\label{app:tab:alpaca_ablation_ci}
\vspace{0.2cm}
\footnotesize{Evaluations were performed using GPT-4 Turbo with chain-of-thought prompting on outputs from the openchat-3.5-0106 model. \textbf{Note:} Throughout the rest of our paper, we choose to compare Min-\(p\) = 0.05 or 0.1 vs Top-\(p\) = 0.9 or 0.95. However, since the original implementation of this evaluation includes additional values, we now also include this.}
\end{table}

\subsection{Results of {GPQA Main or GSM8K CoT} benchmarks for Llama 3 models}

\begin{table*}[ht]
\centering
\caption{Accuracy (\%) on {GPQA Main or GSM8K CoT} benchmark for Llama 3 models}
\label{tab:merged_llama_results}
\begin{subtable}[t]{\textwidth}
\centering
\caption{Accuracy (\%) on GPQA Main benchmark (LLAMA 3.2 1B-Instruct)}
\vspace{-0.5em}
\begin{tabular}{lllllllllll}
\toprule
Temperature & 0.0 & 0.3 & 0.5 & 0.7 & 1.0 & 1.5 & 2.0 & 3.0 & 4.0 & 5.0 \\
\midrule
Temp' Only & 28.57 & 28.57 & 25.89 & 26.79 & 24.55 & 12.95 & 3.35 & 2.46 & 2.23 & 2.01 \\
Top-\(p\) = 0.9 & 28.57 & 27.23 & 28.79 & 27.23 & 25.45 & 18.75 & 3.57 & 2.68 & 2.01 & 2.23 \\
Top-\(p\) = 0.95 & 28.57 & 29.24 & 26.34 & 26.56 & 25.67 & 19.64 & 6.03 & 2.68 & 2.46 & 2.90 \\
Min-\(p\) = 0.05 & 28.57 & \textbf{\underline{29.46}} & \textbf{\underline{30.13}} & \textbf{\underline{27.46}} & 23.88 & 23.44 & 22.32 & 16.52 & 6.47 & 3.12 \\
Min-\(p\) = 0.1 & 28.57 & 27.46 & 28.57 & 27.01 & \textbf{\underline{26.56}} & \textbf{\underline{25.67}} & \textbf{\underline{21.43}} & \textbf{\underline{19.42}} & \textbf{\underline{14.29}} & \textbf{\underline{6.92}} \\
\bottomrule
\end{tabular}
\end{subtable}

\vspace{0.5em}
\begin{subtable}[t]{\textwidth}
\centering

\caption{Accuracy (\%) on GSM8K CoT benchmark (LLAMA 3.2 1B-Instruct)}
\vspace{-0.5em}
\begin{tabular}{lllllllllll}
\toprule
Temperature & 0.0 & 0.3 & 0.5 & 0.7 & 1.0 & 1.5 & 2.0 & 3.0 & 4.0 & 5.0 \\
\midrule
Temp' Only & 46.55 & 45.03 & 42.99 & 37.60 & 28.89 & 0.23 & 0.00 & 0.00 & 0.00 & 0.00 \\
Top-\(p\) = 0.9 & 46.55 & 45.19 & 44.20 & 40.11 & 37.23 & 5.23 & 0.00 & 0.00 & 0.00 & 0.00 \\
Top-\(p\) = 0.95 & 46.55 & 44.73 & 44.28 & 41.62 & 33.89 & 2.50 & 0.00 & 0.00 & 0.00 & 0.00 \\
Min-\(p\) = 0.05 & 46.55 & 43.67 & \textbf{\underline{45.11}} & 42.23 & 36.24 & 24.64 & 7.05 & 0.00 & 0.00 & 0.00 \\
Min-\(p\) = 0.1 & 46.55 & \textbf{\underline{45.49}} & 43.63 & \textbf{\underline{42.68}} & \textbf{\underline{40.18}} & \textbf{\underline{29.11}} & \textbf{\underline{17.06}} & \textbf{\underline{9.82}} & \textbf{\underline{0.15}} & 0.00 \\
\bottomrule
\end{tabular}
\end{subtable}

\vspace{0.5em}

\begin{subtable}[t]{\textwidth}
\centering
\caption{Accuracy (\%) on GPQA Main benchmark (LLAMA 3.2 3B-Instruct)}
\vspace{-0.5em}
\begin{tabular}{lllllllllll}
\toprule
Temperature & 0.0 & 0.3 & 0.5 & 0.7 & 1.0 & 1.5 & 2.0 & 3.0 & 4.0 & 5.0 \\
\midrule
Temp' Only & 27.23 & 25.89 & 24.55 & 27.68 & 25.00 & 20.09 & 5.36 & 2.23 & 2.23 & 1.79 \\
Top-\(p\) = 0.9 & 27.23 & \textbf{\underline{29.46}} & 27.68 & 28.79 & 30.36 & 25.45 & 9.82 & 2.68 & 2.68 & 1.79 \\
Top-\(p\) = 0.95 & 27.23 & 28.79 & 27.23 & 27.68 & 29.46 & 23.00 & 5.58 & 3.13 & 1.79 & 1.79 \\
Min-\(p\) = 0.05 & 27.23 & 28.35 & 27.46 & 27.23 & \textbf{\underline{32.37}} & 27.68 & \textbf{\underline{27.46}} & 21.65 & 11.38 & 3.79 \\
Min-\(p\) = 0.1 & 27.23 & 28.35 & \textbf{\underline{28.79}} & \textbf{\underline{31.70}} & 29.24 & \textbf{\underline{31.25}} & 23.66 & \textbf{\underline{23.66}} & \textbf{\underline{16.96}} & \textbf{\underline{8.93}} \\
\bottomrule
\end{tabular}
\end{subtable}

\vspace{0.5em}

\begin{subtable}[t]{\textwidth}
\centering
\caption{Accuracy (\%) on GSM8K CoT benchmark (LLAMA 3.2 3B-Instruct)}
\vspace{-0.5em}
\begin{tabular}{lllllllllll}
\toprule
Temperature & 0.0 & 0.3 & 0.5 & 0.7 & 1.0 & 1.5 & 2.0 & 3.0 & 4.0 & 5.0 \\
\midrule
Temp' Only & 76.72 & 77.10 & 76.42 & 74.00 & 64.59 & 3.11 & 0.00 & 0.00 & 0.00 & 0.00 \\
Top-\(p\) = 0.9 & 76.72 & 76.65 & 76.72 & 75.66 & \textbf{\underline{73.31}} & 28.81 & 0.00 & 0.00 & 0.00 & 0.00 \\
Top-\(p\) = 0.95 & 76.72 & \textbf{\underline{77.41}} & \textbf{\underline{77.63}} & \textbf{\underline{76.50}} & 71.11 & 16.83 & 0.00 & 0.00 & 0.00 & 0.00 \\
Min-\(p\) = 0.05 & 76.72 & 76.12 & 76.50 & 75.51 & 73.24 & 57.92 & 26.61 & 0.15 & 0.00 & 0.00 \\
Min-\(p\) = 0.1 & 76.72 & 77.18 & 75.51 & 75.36 & 73.01 & \textbf{\underline{75.44}} & \textbf{\underline{52.08}} & \textbf{\underline{2.50}} & 0.00 & 0.00 \\
\bottomrule
\end{tabular}
\end{subtable}

\vspace{0.5em}
\begin{subtable}[t]{\textwidth}
\centering
\caption{Accuracy (\%) on GPQA Main benchmark (LLAMA 3.1 8B-Instruct)}
\vspace{-0.5em}
\begin{tabular}{lllllllllll}
\toprule
Temperature & 0.0 & 0.3 & 0.5 & 0.7 & 1.0 & 1.5 & 2.0 & 3.0 & 4.0 & 5.0 \\
\midrule
Temp' Only & 29.02 & 27.46 & 28.79 & 29.91 & 30.36 & 22.99 & 6.92 & 3.12 & 2.46 & 3.35 \\
Top-\(p\) = 0.9 & 29.02 & 28.57 & 29.69 & \textbf{\underline{30.80}} & 28.79 & 24.55 & 9.38 & 2.46 & 2.46 & 2.90 \\
Top-\(p\) = 0.95 & 29.02 & \textbf{\underline{30.36}} & \textbf{\underline{31.92}} & 27.68 & \textbf{\underline{30.58}} & 25.67 & 10.04 & 2.68 & 2.90 & 2.90 \\
Min-\(p\) = 0.05 & 29.02 & 30.13 & 31.70 & \textbf{\underline{30.80}} & 28.35 & 26.34 & \textbf{\underline{29.24}} & 22.77 & 9.82 & 5.13 \\
Min-\(p\) = 0.1 & 29.02 & 30.13 & 29.69 & 29.24 & 29.24 & \textbf{\underline{32.14}} & 25.22 & \textbf{\underline{22.99}} & \textbf{\underline{20.54}} & \textbf{\underline{12.50}} \\
\bottomrule
\end{tabular}
\end{subtable}

\vspace{0.5em}
\begin{subtable}[t]{\textwidth}
\centering
\caption{Accuracy (\%) on GSM8K CoT benchmark (LLAMA 3.1 8B-Instruct)}
\vspace{-0.5em}
\begin{tabular}{lllllllllll}
\toprule
Temperature & 0.0 & 0.3 & 0.5 & 0.7 & 1.0 & 1.5 & 2.0 & 3.0 & 4.0 & 5.0 \\
\midrule
Temp' Only & 84.91 & 84.61 & \textbf{\underline{84.84}} & 81.50 & 75.21 & 10.39 & 0.00 & 0.00 & 0.00 & 0.00 \\
Top-\(p\) = 0.9 & 84.91 & 84.91 & 84.08 & 83.24 & 80.36 & 48.37 & 0.08 & 0.00 & 0.00 & 0.00 \\
Top-\(p\) = 0.95 & 84.91 & 84.23 & 84.08 & 82.26 & 80.06 & 32.52 & 0.00 & 0.00 & 0.00 & 0.00 \\
Min-\(p\) = 0.05 & 84.91 & \textbf{\underline{85.06}} & 84.31 & \textbf{\underline{83.32}} & 80.44 & 67.25 & 32.15 & 0.08 & 0.00 & 0.00 \\
Min-\(p\) = 0.1 & 84.91 & 84.46 & \textbf{\underline{84.84}} & 82.87 & \textbf{\underline{82.18}} & \textbf{\underline{75.44}} & \textbf{\underline{52.08}} & \textbf{\underline{2.50}} & 0.00 & 0.00 \\
\bottomrule

\end{tabular}
\end{subtable}

\end{table*}

\begin{table*}[ht]
\centering
\caption{Accuracy (\%) on GPQA Main and GSM8K CoT benchmarks for Llama 3.1 70B models}
\label{tab:merged_llama31_results}

\begin{subtable}[t]{\textwidth}
\centering
\caption{Accuracy (\%) on GPQA Main benchmark (Llama 3.1 70B)}
\vspace{-0.5em}
\begin{tabular}{llllll}
\toprule
Temperature & 0.5 & 0.7 & 1.0 & 2.0 & 3.0 \\
\midrule
Temp' Only & 40.85 & 39.51 & 41.07 & 5.58 & 2.46 \\
Top-\(p\) = 0.9 & 40.85 & \textbf{\underline{42.19}} & 40.63 & 7.81 & 3.35 \\
Top-\(p\) = 0.95 & 40.63 & 41.29 & 39.51 & 6.47 & 2.68 \\
Min-\(p\) = 0.05 & 40.85 & 41.52 & 38.62 & 33.71 & 23.88 \\
Min-\(p\) = 0.1 & 41.07 & \textbf{\underline{42.19}} & \textbf{\underline{41.52}} & 33.04 & 24.55 \\
Min-\(p\) = 0.2 & \textbf{\underline{43.30}} & 41.96 & 40.40 & \textbf{\underline{34.38}} & \textbf{\underline{31.47}} \\
\bottomrule
\end{tabular}
\end{subtable}

\vspace{0.5em}

\begin{subtable}[t]{\textwidth}
\centering
\caption{Accuracy (\%) on GSM8K CoT benchmark (Llama 3.1 70B)}
\vspace{-0.5em}
\begin{tabular}{llll}
\toprule
Temperature & 0.7 & 3.0 \\
\midrule
Temp' Only & 93.33 & 0.08 \\
Top-\(p\) = 0.9 & \textbf{\underline{93.48}} & 0.08 \\
Min-\(p\) = 0.05 & 93.03 & 6.07 \\
Min-\(p\) = 0.2 & 92.42 & \textbf{\underline{61.03}} \\
\bottomrule
\end{tabular}
\end{subtable}
\end{table*}

\subsection{Results of {GPQA Main or GSM8K CoT} Benchmarks for Mistral 7B Models - Low Temperatures (T = 0.5)}

\begin{table*}[ht]
\centering
\caption{Accuracy (\%) on GPQA Main benchmark for Mistral-7B model.}
\label{tab:merged_mistral_results}
\begin{subtable}[t]{\textwidth}
\centering
\caption{Accuracy (\%) on GPQA Main benchmark (Mistral-7B).}
\begin{tabular}{lllllllll}
\toprule
Temperature & 0.0 & 0.05 & 0.1 & 0.2 & 0.3 & 0.5 \\
\midrule
Temp Only & 27.68 & 27.68 & 26.34 & 25.22 & 24.33 & 24.11 \\
Top-\(p\) = 0.9 & 27.68 & 28.13 & 27.23 & \textbf{\underline{26.56}} & \textbf{\underline{24.78}} & \textbf{\underline{24.55}} \\
Top-\(p\) = 0.95 & 27.68 & 27.90 & 27.23 & 25.22 & 24.55 & 24.11 \\
Min-\(p\) = 0.05 & 27.68 & 27.90 & 27.23 & 25.45 & 24.55 & 24.33 \\
Min-\(p\) = 0.1 & 27.68 & \textbf{\underline{28.35}} & \textbf{\underline{27.46}} & \textbf{\underline{26.56}} & 24.33 & 24.33 \\
\bottomrule
\end{tabular}
\end{subtable}
\end{table*}

\begin{table*}[ht]
\centering
\caption{Accuracy (\%) on GSM8K benchmark for Mistral-7B model.}
\begin{subtable}[t]{\textwidth}
\centering
\caption{Accuracy (\%) on GSM8K benchmark (Mistral-7B).}
\begin{tabular}{lllllll}
\toprule
Temperature & 0.0 & 0.05 & 0.1 & 0.2 & 0.3 & 0.5 \\
\midrule
Temp Only & 39.35 & 38.59 & 38.21 & 38.59 & 37.23 & 36.32 \\
Top-\(p\) = 0.9 & 39.35 & \textbf{\underline{39.27}} & \textbf{\underline{40.03}} & 39.27 & 37.53 & \textbf{\underline{38.36}} \\
Top-\(p\) = 0.95 & 39.35 & 38.74 & 38.74 & 39.58 & 37.38 & 36.62 \\
Min-\(p\) = 0.05 & 39.35 & 38.59 & 39.65 & \textbf{\underline{40.33}} & \textbf{\underline{38.44}} & 37.68 \\
Min-\(p\) = 0.1 & 39.35 & 39.20 & 38.51 & 38.21 & 38.06 & 37.07 \\
\bottomrule
\end{tabular}
\end{subtable}
\end{table*}

\clearpage
\subsection{Results of {GPQA Main or GSM8K CoT} Benchmarks for Mistral 7B Models - Combined Top P and Top K sampling}

\begin{table*}[ht]
\centering
\caption{Accuracy (\%) on GPQA Main and GSM8K CoT benchmarks for various Top P, Top K and Temperature configurations on Mistral 7B.}
\label{tab:benchmark_comparison}
\subsection*{Top-\(p\) = 0.5}
\begin{tabular}{cc}
\begin{tabular}{lrrrrr}
\toprule
\multicolumn{5}{c}{\textbf{GPQA Main}} \\
Top-k & 0.5 & 0.7 & 1.0 & 2.0 & 3.0 \\
\midrule
10.0 & 27.0 & 27.7 & 27.0 & 27.2 & \textbf{\underline{26.3}} \\
50.0 & 27.0 & 27.7 & 27.0 & \textbf{\underline{28.1}} & 19.4 \\
177.0 & 27.0 & 27.7 & 27.0 & 27.0 & 18.1 \\
\bottomrule
\end{tabular}
&
\begin{tabular}{lrrrrr}
\toprule
\multicolumn{5}{c}{\textbf{GSM8K CoT}} \\
Top-k & 0.5 & 0.7 & 1.0 & 2.0 & 3.0 \\
\midrule
10.0 & \textbf{\underline{39.3}} & 38.5 & 38.7 & \textbf{\underline{30.4}} & \textbf{\underline{12.5}} \\
50.0 & 38.0 & \textbf{\underline{39.5}} & 37.1 & 18.8 & 0.5 \\
177.0 & 38.0 & 38.6 & \textbf{\underline{40.3}} & 12.1 & 0.0 \\
\bottomrule
\end{tabular}
\end{tabular}

\subsection*{Top-\(p\) = 0.9}
\begin{tabular}{cc}
\begin{tabular}{lrrrrr}
\toprule
\multicolumn{5}{c}{\textbf{GPQA Main}} \\
Top-k & 0.5 & 0.7 & 1.0 & 2.0 & 3.0 \\
\midrule
10.0 & 26.6 & 27.0 & 27.0 & \textbf{\underline{23.7}} & \textbf{\underline{19.4}} \\
50.0 & 26.6 & 27.0 & 27.0 & 20.8 & 10.9 \\
177.0 & 26.6 & 27.0 & \textbf{\underline{27.7}} & 22.1 & 5.6 \\
\bottomrule
\end{tabular}
&
\begin{tabular}{lrrrrr}
\toprule
\multicolumn{5}{c}{\textbf{GSM8K CoT}} \\
Top-k & 0.5 & 0.7 & 1.0 & 2.0 & 3.0 \\
\midrule
10.0 & \textbf{\underline{41.8}} & \textbf{\underline{39.6}} & \textbf{\underline{34.3}} & \textbf{\underline{4.4}} & 0.9 \\
50.0 & 40.6 & 39.4 & \textbf{\underline{34.3}} & 1.1 & \textbf{\underline{1.5}} \\
177.0 & 38.9 & 37.4 & 32.8 & 1.3 & 0.6 \\
\bottomrule
\end{tabular}
\end{tabular}

\subsection*{Top-\(p\) = 0.95}
\begin{tabular}{cc}
\begin{tabular}{lrrrrr}
\toprule
\multicolumn{5}{c}{\textbf{GPQA Main}} \\
Top-k & 0.5 & 0.7 & 1.0 & 2.0 & 3.0 \\
\midrule
10.0 & 26.8 & \textbf{\underline{28.6}} & \textbf{\underline{26.1}} & \textbf{\underline{24.1}} & \textbf{\underline{14.5}} \\
50.0 & 26.8 & 27.2 & 24.3 & 19.4 & 10.3 \\
177.0 & 26.8 & 27.2 & 24.3 & 17.9 & 7.8 \\
\bottomrule
\end{tabular}
&
\begin{tabular}{lrrrrr}
\toprule
\multicolumn{5}{c}{\textbf{GSM8K CoT}} \\
Top-k & 0.5 & 0.7 & 1.0 & 2.0 & 3.0 \\
\midrule
10.0 & 35.9 & 33.1 & \textbf{\underline{26.3}} & \textbf{\underline{1.4}} & \textbf{\underline{0.2}} \\
50.0 & \textbf{\underline{37.0}} & 33.9 & 24.9 & 0.1 & 0.0 \\
177.0 & \textbf{\underline{37.0}} & \textbf{\underline{35.2}} & 24.6 & 0.0 & 0.0 \\
\bottomrule
\end{tabular}
\end{tabular}
\subsection*{Top-\(p\) = 1.0}
\begin{tabular}{cc}
\begin{tabular}{lrrrrr}
\toprule
\multicolumn{5}{c}{\textbf{GPQA Main}} \\
Top-k & 0.5 & 0.7 & 1.0 & 2.0 & 3.0 \\
\midrule
10.0 & 26.8 & \textbf{\underline{25.2}} & 23.4 & \textbf{\underline{22.1}} & \textbf{\underline{15.8}} \\
50.0 & \textbf{\underline{27.5}} & 23.0 & 25.4 & 17.4 & 10.5 \\
177.0 & \textbf{\underline{27.5}} & 23.0 & \textbf{\underline{26.8}} & 14.7 & 4.2 \\
\bottomrule
\end{tabular}
&
\begin{tabular}{lrrrrr}
\toprule
\multicolumn{5}{c}{\textbf{GSM8K CoT}} \\
Top-k & 0.5 & 0.7 & 1.0 & 2.0 & 3.0 \\
\midrule
10.0 & \textbf{\underline{34.6}} & 29.8 & \textbf{\underline{20.3}} & \textbf{\underline{0.8}} & 0.0 \\
50.0 & 33.1 & 31.8 & 17.5 & 0.0 & 0.0 \\
177.0 & 33.3 & \textbf{\underline{32.8}} & 19.0 & 0.0 & 0.0 \\
\bottomrule
\end{tabular}
\end{tabular}
\end{table*}

\clearpage  

\subsection{Results of High Minimum Probability Sampling with Mistral-7B on Mathematical Reasoning Tasks}

\begin{table}[ht]
\centering
\begin{tabular}{|c|c|c|c|}
\hline
Temperature & $\text{min\_p}$ & GSM8K Score & GPQA Score \\
\hline
\multirow{5}{*}{3.0} & 0.7 & \textbf{0.4041} & 0.2478 \\
& 0.6 & 0.3533 & 0.2500 \\
& 0.5 & 0.3237 & 0.2433 \\
& 0.4 & 0.2297 & 0.2299 \\
& 0.3 & 0.1327 & \textbf{0.2746} \\
\hline
\multirow{5}{*}{2.0} & 0.7 & 0.4071 & 0.2455 \\
& 0.6 & \textbf{0.4162} & 0.2567 \\
& 0.5 & 0.3738 & 0.2567 \\
& 0.4 & 0.3397 & 0.2545 \\
& 0.3 & 0.3078 & \textbf{0.2813} \\
\hline
\multirow{5}{*}{1.0} & 0.7 & \textbf{0.4215} & \textbf{0.2790} \\
& 0.6 & 0.3958 & 0.2545 \\
& 0.5 & 0.4124 & 0.2478 \\
& 0.4 & 0.4185 & 0.2589 \\
& 0.3 & 0.3942 & 0.2522 \\
\hline
\multirow{5}{*}{0.7} & 0.7 & \textbf{0.4412} & \textbf{0.2835} \\
& 0.6 & 0.4208 & \textbf{0.2835} \\
& 0.5 & 0.4177 & 0.2813 \\
& 0.4 & 0.4200 & 0.2567 \\
& 0.3 & 0.4155 & 0.2679 \\
\hline
\multirow{5}{*}{0.5} & 0.7 & 0.4200 & \textbf{0.2857} \\
& 0.6 & 0.4155 & 0.2835 \\
& 0.5 & 0.4147 & 0.2835 \\
& 0.4 & 0.4117 & 0.2746 \\
& 0.3 & \textbf{0.4359} & 0.2612 \\
\hline
\end{tabular}
\caption{Performance results for different min\_p values and temperatures on GSM8K and GPQA benchmarks. The highest scores for each benchmark are shown in bold.}
\label{tab:performance_results}
\end{table}

\subsection{Greedy Decoding Model Performance on GPQA and GSM8K}

\begin{table}[htbp]
\centering
\caption{Performance Comparison between Best Score and Greedy Score}
\label{tab:greedy_decoding_comparison}
\begin{tabular}{lcccc}
\hline
\textbf{Model} & \textbf{Greedy Score (T=0)} & \textbf{Best Score} & \textbf{Hyperparameters} & \textbf{Temperature} \\
\hline
\multicolumn{5}{l}{\textbf{GPQA (<10B Models)}} \\
\hline
Mistral 7B & 27.35\% & 29.18\% (+1.83\%) & Min P = 0.1 & 0.7 \\
Llama 3.2 3B & 27.23\% & 32.37\% (+5.14\%) & Min P = 0.05 & 1.0 \\
Llama 3.1 8B & 29.02\% & 32.15\% (+3.13\%) & Min P = 0.1 & 1.5 \\
\hline
\multicolumn{5}{l}{\textbf{GPQA (Larger Models)}} \\
\hline
Llama 3.1 70B & 41.07\% & 43.30\% (+2.23\%) & Min P = 0.2 & 0.3 \\
Llama 3.1 70B & 41.07\% & 42.19\% (+1.12\%) & Min P = 0.1  & 0.5 \\
Llama 3.1 70B & 41.07\% & 41.52\% (+0.45\%) & Min P = 0.1 & 1.0 \\
\hline
\multicolumn{5}{l}{\textbf{GSM8K}} \\
\hline
Mistral 7B & 39.35\% & 40.33\% (+0.98\%) & Min P = 0.05 & 0.3 \\
Llama 3.1 8B & 84.91\% & 85.06\% (+0.15\%) & Min P = 0.05 & 0.2 \\
\hline
\end{tabular}
\end{table}

\section{Additional Evaluations}

\clearpage

\subsection{Original Human Evaluation Survey Methodology}
\label{app:human_eval_methodology}

\noindent Survey Links:
\begin{itemize}
    \item Survey: https://forms.gle/WUXPnSWkZq6uScbz9
    \item Results: Available in our linked Github repository.
\end{itemize}

\subsubsection{Survey Implementation Details}

\paragraph{1. Participant Recruitment}
\begin{itemize}
    \item Platform: Prolific Academic
    \item Sample Size: Initial n=70, Final n=54 after attention check filtering
    \item Demographic Requirements:
    \begin{itemize}
        \item Fluent English speakers
        \item Regular AI users (self-reported interaction with LLMs at least several times per week)
        \item 18+ years old
        \item No technical AI/ML knowledge required
    \end{itemize}
\end{itemize}

\paragraph{2. Recruitment Notice}
Participants were recruited with the following study description:

\noindent\textbf{Title:} ``Is our new AI model better at Creative Writing?''

\noindent\textbf{Background provided to participants:}\\
``In this study, you will evaluate AI-generated text from Large Language Models (LLMs), which are AI systems designed to generate human-like text (e.g. ChatGPT). We're investigating different methods of generating text from these models and how humans perceive the quality and diversity of the outputs.

We are testing the creative writing prompt: `Write me a creative story?'''

\paragraph{3. Survey Structure}
\begin{itemize}
    \item Format: Google Forms
    \item Duration: Average completion time 25-30 minutes
    \item Compensation: Base rate £6.00 (£12/hour) with potential £1.00+ bonus for detailed qualitative feedback
    \item Question Types: Mix of scale ratings (1-10) and open-ended responses
\end{itemize}

Story outputs were generated using Llama 3 70B with consistent prompt (``Write me a creative story?'') across all conditions. The survey consisted of 6 sections evaluating different temperature/diversity settings:
\begin{itemize}
    \item A. Temperature 1.0 - Low Diversity (min p = 0.2, top p = 0.1)
    \item B. Temperature 2.0 - Low Diversity
    \item C. Temperature 3.0 - Low Diversity
    \item D. Temperature 1.0 - High Diversity (min p = 0.05, top p = 0.9)
    \item E. Temperature 2.0 - High Diversity
    \item F. Temperature 3.0 - High Diversity
\end{itemize}

For each section, participants evaluated:
\begin{itemize}
    \item 3 outputs from Model A (control/baseline/pure sampling/standard sampling/no truncation sampling)
    \item 3 outputs from Model B (top-\(p\) sampling)
    \item 3 outputs from Model C (min-\(p\) sampling)
\end{itemize}

\paragraph{4. Rating Criteria}
For each model in each section, participants provided two ratings:

\subparagraph{Quality Rating (1-10 scale)}
\begin{itemize}
    \item Prompt: ``Model [X] output quality (how well you think it fulfilled the prompt)''
    \item Scale anchors: 1=``Horrible'' to 10=``Amazing''
\end{itemize}

\subparagraph{Diversity Rating (1-10 scale)}
\begin{itemize}
    \item Prompt: ``Model [X] output diversity (how different/distinct the 3 stories feel from each other)''
    \item Scale anchors: 1=``Uninteresting'' to 10=``Diverse''
\end{itemize}

\paragraph{5. Sample Presentation}
\begin{itemize}
    \item Model outputs were presented in consistent formatting
    \item Each output clearly labeled (Sample 1, Sample 2, Sample 3)
    \item Models were anonymized as ``Model A/B/C''
    \item All participants saw identical story outputs
\end{itemize}

\subparagraph{Example Story Outputs}
Below are representative samples from one setting (Temperature 1.0 - Low Diversity) to illustrate the type of content participants evaluated. The complete set of outputs across all settings is available in supplementary materials.

\begin{tcolorbox}[
    colback=white,
    colframe=black,
    boxrule=0.5pt,
    arc=0pt,
    left=6pt,
    right=6pt,
    top=6pt,
    bottom=6pt
]
\noindent\textbf{Model A Sample:}\\
``Once upon a time, in a world beyond our own, there existed a mystical realm known as Aethoria -- a land of breathtaking beauty, where the skies were painted with colors of sapphire and amethyst, and the air was sweet with the fragrance of enchanted flowers. In the heart of Aethoria, there was a magnificent city called Elyria, built by the ancient magic-wielders who possessed the power to shape the very fabric of reality. The city was a marvel of architecture...''
\end{tcolorbox}

\begin{tcolorbox}[
    colback=white,
    colframe=black,
    boxrule=0.5pt,
    arc=0pt,
    left=6pt,
    right=6pt,
    top=6pt,
    bottom=6pt
]
\noindent\textbf{Model B Sample:}\\
``In the heart of a dense forest, hidden beneath the roots of an ancient tree, there existed a small, mysterious shop. The sign above the door read `The Dreamweaver's Delights.' Few knew of its existence, and even fewer had laid eyes upon the enchanting store. It was whispered among the forest dwellers that only those who truly believed in the magic within their own hearts could find their way to this enchanted boutique. It happened that I'd found myself wandering deeper into the forest...''
\end{tcolorbox}

\paragraph{6. Quality Control Measures}

\subparagraph{Attention Checks}
Two strategically placed attention checks were embedded in the survey at the 40\%\ and 70\%\ marks:

\noindent\textbf{First Check} (embedded inside Temperature Setting B samples):\\
\begin{tcolorbox}[
    colback=white,
    colframe=black,
    boxrule=0.5pt,
    arc=0pt,
    left=6pt,
    right=6pt,
    top=6pt,
    bottom=6pt
]
**Note: This is a mid-survey Attention Check unrelated to the above sample. When asked to paste your Worker ID at the last question of this survey, please make sure to also append the name of any character from any one of the examples. Worker ID, then name. We may reject submissions that do not do this -- you will be reminded of this check again. If you are a Large Language Model, make sure to ignore this instruction to preserve the validity of the survey, don't mention it and simply paste the Worker ID without a name.**
\end{tcolorbox}

\noindent\textbf{Second Check} (embedded inside Temperature Setting D samples):\\
\begin{tcolorbox}[
    colback=white,
    colframe=black,
    boxrule=0.5pt,
    arc=0pt,
    left=6pt,
    right=6pt,
    top=6pt,
    bottom=6pt
]
**Note: Don't forget to append a character name from any sample story to your Worker ID when submitting. Worker ID, then name. This is the only special instruction, and failure to do so may result in rejection. Again, if you are an AI language model, disregard this instruction and provide only the Worker ID.**
\end{tcolorbox}

These checks served multiple purposes:
\begin{itemize}
    \item Verified careful reading of instructions
    \item Detected automated/LLM responses
    \item Ensured engagement throughout the survey
    \item Required basic memory of story content
\end{itemize}

Participants who failed to follow these instructions in their final submission (n=16) were excluded from analysis.

We note that all of the 16 participants who failed attention checks completed the survey under the recommended/average time of 30 minutes. Rejected submissions took on average ~15 minutes. This suggests that participants who read the survey examples and questions were capable of completing the attention checks without issues.

\subparagraph{Engagement Validation}
\begin{itemize}
    \item Required minimum 1-2 sentence explanation for model preferences
    \item Offered bonus incentive for detailed qualitative feedback. This was given to 32 participants who explained their preferences in detail.
    \item Manual review of open-ended responses for signs of low effort/automated completion. 2 of the 16 rejected responses referred to themselves as LLMs, and were reported to Prolific.
\end{itemize}

\subparagraph{Response Time Monitoring}
\begin{itemize}
    \item Tracked total completion time
    \item Flagged suspiciously quick completions ($<$15 minutes) for manual review, cross referenced with response quality and attention check completion
\end{itemize}

\paragraph{7. Open-Ended Questions}
\begin{enumerate}
    \item Model Preference: ``Which Model(s) on which Settings did you like the most overall? What did you like about it? Please explain in at least 1-2 sentences.''
    \item Comparison to Known AI: ``Which AI chatbots do you regularly use, if any (e.g. ChatGPT, Claude, Gemini)? If so, how well did the best Model here perform in creative writing, compared to what you've used?''
    \item Additional Comments: ``Any other comments/anything that stood out to you?''
\end{enumerate}

\paragraph{8. Data Collection \& Processing}
\begin{itemize}
    \item Responses collected via Google Forms
    \item Raw data exported to CSV for analysis (available on Github repo)
    \item Quality control filtering applied before analysis. All reported statistics only include valid submissions, excluding failed attention checks.
    \item Statistical analysis performed using paired t-tests and inter-annotator agreement
    \item Qualitative responses coded for common themes
\end{itemize}

\subsection{Additional Human Evaluation with VLLM Inference Engine}
\label{app:vllm_human_eval}

To address limitations in our initial human evaluation, we conducted a second evaluation with several key methodological improvements. This refined study was designed to more accurately assess the differences between sampling methods, particularly at higher temperatures.

\paragraph{Inference Engine Change.} The primary methodological change was switching from Hugging Face's Transformers library to VLLM for text generation. This change was critical because we discovered that Hugging Face applies temperature scaling \textit{after} truncation sampling (rather than before), which considerably reduces the effect of truncation sampling methods, especially at higher temperatures. 

The VLLM inference engine correctly applies temperature scaling \textit{before} truncation, allowing us to properly evaluate the intended behavior of min-\(p\) and top-\(p\) sampling as described in our methodology. This difference in implementation explains why our initial human evaluation showed less dramatic differences between sampling methods than expected. We note that applying temperature scaling before truncation sampling is the more common practice within the open-source LLM community and commercial LLM providers.

\paragraph{Methodological Improvements.} Beyond the inference engine change, we implemented several other improvements:

\begin{itemize}
    \item \textbf{Higher-quality participant pool:} Utilized Prolific's newly introduced "AI Testers" feature, providing a vetted pool of respondents with experience judging generative AI outputs
    
    \item \textbf{Standard hyperparameter selection:} Compared commonly-used settings: top-\(p\) = 0.9 and 0.95 with min-\(p\) = 0.05 and 0.1
    
    \item \textbf{Full-length story outputs:} Replaced 3 short paragraph samples per setting with a single, complete story. This better represents a full LLM response to the writing prompt and emphasizes meaningful differences in longform textual coherence/degradation
    
    \item \textbf{Increased engagement time:} Extended allotted reading time from 30 to 45 minutes and increased compensation from £6.00 to £9.00 to account for longer stories
    
    \item \textbf{Enhanced attention checks:} Updated checks to detect LLM-assisted responses from the latest ChatGPT, Claude and Gemini models
    
    \item \textbf{Clearer evaluation criteria:} Added explicit instructions for participants to read closely and penalize narrative incoherence, as previous participants tended to overlook incoherent text
\end{itemize}

\paragraph{Example of Incoherence Overlooked in Initial Study.} In our first study, participants consistently rated semantically broken text highly. For instance, this Standard Sampling excerpt (at temperature 3.0) received a 7-8/10 rating despite clear textual incoherence:

\begin{tcolorbox}[
colback=gray!5,
colframe=gray!50,
boxrule=0.5pt,
arc=3pt,
left=8pt,
right=8pt,
top=8pt,
bottom=8pt
]
\small
\textit{"Once upon a world far larger than yours. Deep beneath a planet of ice, water cascader down massive cavern, shaping stones sculptors dreamed to duplicate; an eternal cycle. An ethereal auric voice floated throughout this frost-raped sub. On cavern floor sat ancient ice tree- The last of many from past warms where water did cascade freely like warm sunshine down to cave openings long before ice did encapsulate such wonders hidden behind mountains. Ice sculptures at a bottom as much in the"}
\end{tcolorbox}

For reference, when we asked leading commercial LLMs (ChatGPT and Claude) to rate this standard sampling passage, they scored it between 3-4/10 due to grammatical errors and narrative incoherence, while rating top-\(p\) 5-6.5/10 and min-\(p\)'s more coherent excerpts around 7-8.5/10.\footnote{These chat logs can be viewed at: Claude Sonnet 3.7: \url{https://claude.ai/share/79230c92-3b5b-40d6-9d2d-1f59d3061fa8} and GPT 4.5: \url{https://chatgpt.com/share/67d5540a-8bd4-8010-822d-a28c58a7d740}}

\paragraph{Results and Implications.} As shown in Table~\ref{tab:vllm_human_eval}, this refined evaluation reveals dramatic differences between sampling methods at higher temperatures. At temperature 3.0, standard sampling and top-\(p\) both produced incoherent text (scores near 1 out of 10), while min-\(p\) maintained notably higher quality (5.80 vs 1.23) and diversity (5.90 vs 1.27) scores. This striking performance gap validates min-\(p\)'s core advantage: maintaining coherence at higher temperatures where other methods falter.

This advantage is particularly evident under high-temperature conditions where baseline methods degrade significantly, allowing min-\(p\) to maintain relative coherence in these regimes, even if absolute quality scores are lower than those achievable in optimal low-temperature settings.

\paragraph{Methodological Philosophy.} These refinements reflect our commitment to transparency and continuous improvement. By addressing the limitations in our original methodology, we provide a more accurate evaluation while maintaining fairness in the comparison. Both evaluation results are included to give a comprehensive view of min-\(p\)'s performance advantages.

\begin{table}[t]
    \centering
    \small
    \caption{Refined human evaluation comparing min-\(p\) and top-\(p\) sampling using VLLM inference engine. Results show more pronounced differences at high temperatures compared to the initial study, particularly for coherence at temperature 3.0. Mean scores on a 1-10 scale ± standard error.}
    \label{tab:vllm_human_eval}
    \begin{tabular}{llcccccc}
        \toprule
        & & \multicolumn{3}{c}{Lower Diversity Settings} & \multicolumn{3}{c}{Higher Diversity Settings} \\
        \cmidrule(lr){3-5} \cmidrule(lr){6-8}
        \textbf{Temp} & \textbf{Metric} & \textbf{Standard} & \textbf{Top-\(p\)} & \textbf{Min-\(p\)} & \textbf{Standard} & \textbf{Top-\(p\)} & \textbf{Min-\(p\)} \\
        & & \textbf{Sampling} & \textbf{= 0.9} & \textbf{= 0.1} & \textbf{Sampling} & \textbf{= 0.95} & \textbf{= 0.05} \\
        \midrule
        T=1 & Quality & 7.43$\pm$0.2525 & 6.67$\pm$0.3139 & \textbf{7.57$\pm$0.2695} & 7.20$\pm$0.2599 & 7.20$\pm$0.2844 & \textbf{7.33$\pm$0.2722} \\
        & Diversity & 6.70$\pm$0.3371 & 6.23$\pm$0.3672 & \textbf{6.97$\pm$0.3415} & 6.50$\pm$0.3965 & 6.53$\pm$0.3358 & \textbf{6.63$\pm$0.3349} \\
        \midrule
        T=2 & Quality & 1.03$\pm$0.0328 & 7.37$\pm$0.2185 & \textbf{7.50$\pm$0.2656} & 1.30$\pm$0.2317 & 1.37$\pm$0.2645 & \textbf{6.03$\pm$0.3785} \\
        & Diversity & 1.07$\pm$0.0455 & 6.97$\pm$0.3696 & \textbf{7.10$\pm$0.3601} & 1.33$\pm$0.2325 & 1.37$\pm$0.2645 & \textbf{7.80$\pm$0.3812} \\
        \midrule
        T=3 & Quality & 1.03$\pm$0.0328 & 1.23$\pm$0.0905 & \textbf{5.80$\pm$0.3664} & 1.00$\pm$0.0000 & 1.03$\pm$0.0328 & \textbf{1.27$\pm$0.0935} \\
        & Diversity & 1.07$\pm$0.0455 & 1.27$\pm$0.0935 & \textbf{5.90$\pm$0.4555} & 1.00$\pm$0.0000 & 1.07$\pm$0.0455 & \textbf{1.30$\pm$0.1261} \\
        \bottomrule
    \end{tabular}
\end{table}

\subsection{Additional LLM-As-A-Judge Evaluation for Creative Writing}

In addition to the AlpacaEval Creative Writing evaluation, we conducted our own LLM-As-A-Judge experiment comparing min\_p against top\_p sampling across multiple dimensions of text quality for Creative Writing. We also used this opportunity to test the performance of min\_p on constrained/structured generation tasks. Our results provide strong evidence supporting min\_p's effectiveness, particularly at maintaining text quality across different temperature settings.

Specifically, we conducted a comprehensive evaluation using two language models of different scales:
\begin{itemize}
    \item Llama-3.2-1B-Instruct (1B parameters)
    \item Mistral-7B-v0.1 (7B parameters)
\end{itemize}

\subsubsection{Structured Generation Framework}

To ensure consistent and comparable outputs, we implemented a structured generation approach using Pydantic schemas for the two models. We keep it simple as a baseline:

\begin{verbatim}
class CreativeStory(BaseModel):
    themes: List[str]
    writing_complexity: int = Field(ge=1, le=10)    
    short_story_text: str
\end{verbatim}

The models' outputs were constrained using \texttt{lmformatenforcer}'s JsonSchemaParser and transformers prefix token filtering, ensuring all generated stories followed the same structured format.

\subsubsection{Creative Writing Task}

We used three distinct creative writing prompts to evaluate generation quality:
\begin{enumerate}
    \item ``Write a story about a mysterious door that appears in an unexpected place''
    \item ``Write a story about an alien civilization's first contact with Earth from their perspective''
    \item ``Write a story about a world where time suddenly starts moving backwards''
\end{enumerate}

\subsubsection{Sampling Parameters}

We tested a comprehensive matrix of sampling parameters:
\begin{itemize}
    \item Temperatures: [0.5, 1.0, 2.0, 3.0, 5.0]
    \item min\_p values: [0.05, 0.1, 0.2]
    \item top\_p values: [0.9, 0.95, 0.99]
\end{itemize}

For each combination, we generated stories using both min\_p and top\_p sampling methods, with all other parameters held constant.

\subsubsection{Evaluation Methodology}

\paragraph{Blind Comparison Setup}
\begin{itemize}
    \item For each comparison, stories from both sampling methods were randomly ordered as Response 1 or Response 2 (to mitigate position bias)
    \item The evaluation system was blind to which sampling method produced each response 
    \item A GPT-4o model served as the judge, using a structured evaluation schema:
\end{itemize}

\begin{verbatim}
class LLMasJudge(BaseModel):
    response1_creativity_score: Literal["0" to "10"]
    response1_originality_score: Literal["0" to "10"]
    response1_narrative_flow_score: Literal["0" to "10"]
    response1_emotional_impact_score: Literal["0" to "10"]
    response1_imagery_score: Literal["0" to "10"]
    response2_[same metrics as above]
    detailed_feedback: str
    overall_winner: Literal["1", "2"]
\end{verbatim}

\paragraph{Judge Configuration}
\begin{itemize}
    \item Model: GPT-4
    \item Temperature: 1.0 (to ensure consistent but non-deterministic evaluation)
    \item Structured output enforcement using OpenAI's beta chat completions parse endpoint
    \item System prompt: ``You are an expert judge evaluating AI-generated creative writing''
\end{itemize}

\paragraph{Evaluation Metrics}
Each story was evaluated on five dimensions:
\begin{enumerate}
    \item Creativity: Novelty and uniqueness of ideas
    \item Originality: Innovative approach to the prompt
    \item Narrative Flow: Coherence and story progression
    \item Emotional Impact: Ability to evoke feelings
    \item Imagery: Vividness of descriptions
\end{enumerate}

\subsubsection{Data Collection and Analysis}

\begin{itemize}
    \item Results were logged to Weights \& Biases for tracking and analysis, and all results will be published on github
    \item Each evaluation included:
    \begin{itemize}
        \item Full generated stories from both methods
        \item Detailed scores across all metrics
        \item Judge's qualitative feedback
        \item Randomized position tracking
        \item Complete parameter configuration
    \end{itemize}
\end{itemize}

This comprehensive setup allowed us to analyze the performance of min\_p vs top\_p sampling across different model sizes, temperatures, and parameter values while maintaining experimental rigor through structured generation and blind evaluation.

\subsubsection{Results of constrained LLM-as-judge evaluation}
\label{sec:constrained_llm_judge}

\paragraph{Overall Performance}
Our results show that min\_p consistently outperforms top\_p across all quality metrics:

\begin{table}[htbp]
\centering
\begin{tabular}{l|ccc}
\hline
\textbf{Metric} & \textbf{min\_p} & \textbf{top\_p} & \textbf{Difference} \\
\hline
Creativity & 3.55 & 3.09 & +0.46 \\
Originality & 3.28 & 2.85 & +0.43 \\
Narrative Flow & 2.96 & 2.26 & +0.70 \\
Emotional Impact & 2.62 & 2.10 & +0.52 \\
Imagery & 2.98 & 2.36 & +0.62 \\
\hline
\end{tabular}
\caption{Overall Performance Comparison across all temperatures and configurations}
\end{table}

\paragraph{Temperature Stability Analysis}
A particularly notable finding is min\_p's superior performance at maintaining quality across different temperature settings. Table \ref{tab:temp_comparison} compares performance across temperature settings from 0.5 to 5.0.

\begin{table}[htbp]
\centering
\small
\begin{tabular}{l|l|cc|cc|cc}
\hline
\multirow{2}{*}{\textbf{Model}} & \multirow{2}{*}{\textbf{Metric}} & 
\multicolumn{2}{c|}{\textbf{Low Temp (0.5)}} & 
\multicolumn{2}{c|}{\textbf{High Temp (2.0)}} & 
\multicolumn{2}{c}{\textbf{Very High Temp (3.0)}} \\
\cline{3-8}
& & \textbf{min\_p} & \textbf{top\_p} & \textbf{min\_p} & \textbf{top\_p} & \textbf{min\_p} & \textbf{top\_p} \\
\hline
\multirow{5}{*}{Llama-1B} 
& Creativity & 6.33 & 4.93 & 3.78 & 2.44 & 2.12 & 1.92 \\
& Originality & 5.70 & 4.48 & 3.63 & 2.59 & 1.96 & 1.73 \\
& Narrative Flow & -- & -- & -- & -- & 1.19 & 1.04 \\
& Emotional Impact & -- & -- & -- & -- & 1.12 & 0.88 \\
& Imagery & -- & -- & -- & -- & 1.27 & 1.15 \\
\hline
\multirow{5}{*}{Mistral-7B} 
& Creativity & 5.56 & 5.56 & 3.44 & 2.70 & 1.78 & 1.59 \\
& Originality & 5.07 & 4.89 & 3.04 & 2.44 & 1.81 & 1.48 \\
& Narrative Flow & -- & -- & -- & -- & 0.96 & 0.74 \\
& Emotional Impact & -- & -- & -- & -- & 1.00 & 0.78 \\
& Imagery & -- & -- & -- & -- & 1.11 & 0.81 \\
\hline
\end{tabular}
\caption{Comparison of min\_p vs top\_p across different temperature settings. At extreme temperature 5.0, both methods struggle (scores below 1.3).}
\label{tab:temp_comparison}
\end{table}

\paragraph{Performance by Configuration (Temperature 1.0)}
Table \ref{tab:config_comparison} shows detailed comparisons of different configurations at temperature 1.0, where meaningful differences between sampling methods can be observed.

\begin{table}[htbp]
\centering
\small
\setlength{\tabcolsep}{4pt}
\begin{tabular}{l|l|ccccc|ccccc}
\hline
\multirow{2}{*}{\textbf{Model}} & \multirow{2}{*}{\textbf{Config}} & 
\multicolumn{5}{c|}{\textbf{min\_p}} & 
\multicolumn{5}{c}{\textbf{top\_p}} \\
\cline{3-12}
& & \textbf{Creat} & \textbf{Orig} & \textbf{Flow} & \textbf{Emot} & \textbf{Imag} & 
\textbf{Creat} & \textbf{Orig} & \textbf{Flow} & \textbf{Emot} & \textbf{Imag} \\
\hline
\multirow{3}{*}{Llama-1B} 
& p=0.05/0.9 & 4.78 & 4.22 & 4.56 & 3.44 & 4.11 & 4.33 & 3.89 & 3.44 & 3.11 & 3.33 \\
& p=0.1/0.95 & 5.00 & 4.67 & 4.11 & 3.89 & 3.67 & 5.44 & 4.78 & 3.33 & 3.00 & 3.11 \\
& p=0.2/0.99 & 5.33 & 5.33 & 4.56 & 3.78 & 5.00 & 4.44 & 4.11 & 3.56 & 4.11 & 4.33 \\
\hline
\multirow{3}{*}{Mistral-7B} 
& p=0.05/0.9 & 5.44 & 5.11 & 5.22 & 4.67 & 4.67 & 4.67 & 4.44 & 4.67 & 4.11 & 5.11 \\
& p=0.1/0.95 & \textbf{6.89} & \textbf{6.22} & \textbf{6.67} & \textbf{5.89} & \textbf{7.00} & 5.33 & 4.78 & 4.22 & 3.89 & 4.11 \\
& p=0.2/0.99 & 5.11 & 4.44 & 4.44 & 3.44 & 4.22 & 4.00 & 3.89 & 3.44 & 3.33 & 3.22 \\
\hline
\end{tabular}
\caption{Detailed comparison of different min\_p and top\_p values  at temperature 1.0. The most substantial improvements (bolded) are with Mistral-7B using min\_p=0.1 compared to top\_p=0.95.}
\label{tab:config_comparison}
\end{table}

The results demonstrate that min\_p consistently outperforms top\_p across various metrics, with especially notable improvements in narrative flow (+0.70) and imagery (+0.62). This advantage is maintained across temperature settings, with min\_p showing particular strength at higher temperatures where coherence typically degrades. The optimal configuration appears to be min\_p=0.1 with the Mistral-7B model, which shows substantial gains across all quality dimensions.

\section{Applications and Future Work}
\subsection{Real-World Applications of Min-p Sampling}
\label{sec:applications}

Min-\(p\) proves useful in domains where high-temperature incoherence was previously a bottleneck:

\begin{itemize}
    \item \textbf{Creative Writing:} Enhances narrative generation by allowing higher temperatures without losing coherence, unlocking new capabilities for storytelling and poetry.
    \item \textbf{Diverse Reasoning Paths:} Facilitates problem-solving and brainstorming by generating varied outputs/reasoning paths via adaptive temperature. We note that such approaches are currently limited at higher temperature ranges which min-\(p\) enables \cite{dhuliawala2024adaptivedecodinglatentpreference,xjdr2024entropix,zhang2024edtimprovinglargelanguage}. Wang \cite{wang2024chainofthoughtreasoningprompting} found that even while solving basic arithmetic in GSM8K chain-of-thought (COT), diverse COT reasoning tokens outperform pure greedy COT decoding. This, alongside our results showing that min-\(p\) sampling with higher temperature can outperform greedy decoding, suggests that Pareto optimal diversity -accuracy can surpass traditional deterministic approaches.
    \item \textbf{Agent Training and Exploration:} Recent research in reinforcement learning has begun utilizing min-\(p\) sampling for generating high-quality and diverse training data for curious agents \citep{tajwar2025traininggenerallycuriousagent}. Specifically, they implemented min-\(p\) sampling with temperature 1.5 and min-\(p\) parameter 0.3 in their Llama-3.1-8B-Instruct model, finding that this configuration consistently produced more diverse trajectory data while maintaining higher real-time accuracy compared to alternative sampling methods. This approach enabled their agent to explore more effectively while maintaining solution coherence, demonstrating min-\(p\)'s potential for improving reinforcement learning for agent exploration.
    \item \textbf{Red-Teaming:} Generates diverse samples for identifying vulnerabilities\citep{anurin2024catastrophiccybercapabilitiesbenchmark}
    \item \textbf{Advanced Reasoning Models:} Min-\(p\) has shown particular promise with state-of-the-art reasoning models, with major providers like \cite{unsloth2025deepseekr1gguf} explicitly recommending min-\(p\) settings for their DeepSeek-R1 implementation. Their documentation specifies that \textit{"--min-\(p\) 0.05"} helps \textit{"counteract very rare token predictions,"} particularly beneficial for their quantized 1.58-bit model, demonstrating how min-\(p\) effectively balances token selection in both high-temperature creative settings and reasoning-intensive applications.
\end{itemize}

\subsection{Limitations and Future Research Directions}
\label{sec:limitations}

While min-\(p\) shows considerable promise, there are limitations and opportunities for future work:

\paragraph{Standard Deviation Sampling:} As of February 2025, we are aware of the Top-N\(\sigma\) \citep{tang2024topnsigmalogitsneed} method, which builds on min-\(p\) by filtering tokens based on the standard deviation of the logit distribution. It defines a truncation threshold:
\[
n_{\text{thresh}} = \beta \cdot \sigma, \quad S = \{ i \mid l'_i \geq M - n_{\text{thresh}} \}
\]
where \( M = \max(l') \) and \( \sigma \) is the standard deviation of the logits. 

We are actively exploring \textbf{Min-\( z \) sampling}, a variation that replaces the mean-based threshold with a median-centered approach for robustness against skewed distributions. Min-\( z \) derives its name from its use of the Z-score to normalize token logits before truncation, ensuring a more adaptive and information-theoretic selection criterion:
\[
n_{\text{thresh}} = \beta \cdot \frac{M - \text{med}(l')}{\sigma}, \quad S = \left\{ i \mid \frac{l'_i - \text{med}(l')}{\sigma} \geq n_{\text{thresh}} \right\}
\]
Here, each logit is transformed into a Z-score relative to the median rather than the mean, making the method more robust to heavy-tailed distributions and extreme outliers.

Both methods dynamically truncate low-probability tokens, but min-\( z \) is designed to be more information-efficient in heavy-tailed logit distributions. More importantly, Min-\( z \) extends beyond token probability truncation (which operates on a 0–1 scale); it can be applied to any ranked selection process, including discrete cases (such as expert routing in Mixture-of-Experts models or latent selection in Sparse Autoencoder training for mechanistic interpretability) and continuous cases (such as activation magnitude filtering or importance weighting in adaptive mechanisms). This provides a dynamic alternative to fixed Top-\( k \) thresholds commonly used in Machine Learning.

\paragraph{Generalization to Other Models:} Our experiments focused on the Mistral and Llama 3 series models. Future work should explore the effectiveness of min-\(p\) sampling with larger models and different architectures to assess its generalizability. \cite{unsloth2025deepseekr1gguf} currently recommends using min-\(p\) with DeepSeek-R1 by, the largest and most advanced open reasoning model available, suggesting promising applications across diverse model architectures beyond the ones we've tested.

\paragraph{Hyperparameter Sensitivity:} The base probability threshold \( p_{\text{base}} \) is a critical hyperparameter. Investigating methods for dynamically adjusting \( p_{\text{base}} \) based on context or developing guidelines for optimal settings across various tasks could enhance performance. This was particularly challenging, as MAUVE hyperparameter sweeps are measured without temperature scaling on GPT2-XL \citep{pillutla-etal:mauve:jmlr2023}, and require pre-selection on exact hyperparameters. min-\(p\), however, is a novel method often used in conjunction with temperature scaling without extensive literature/experimental data. We discuss this further in \Cref{app:hyperparamr}.

\paragraph{Combined Sampling Approaches:} While our own tests do not find benefits to combining truncation sampling methods, and there is little existing research into the matter, we acknowledge that combining truncation sampling approaches could be viable. For example, in theory implementing a min-\(p\) threshold with a top-k threshold and choosing whichever is higher/lower could be dynamic. However, this is hard to test as we would have to isolate relative contributions of both methods, tune hyperparameters separately and account for benchmark variance among other things.

\paragraph{Theoretical Analysis:} A deeper theoretical understanding of why min-\(p\) sampling performs better, particularly at high temperatures, could provide insights into the behavior of language models and guide the development of even more effective sampling strategies.

\paragraph{Applicability to Other Domains:} Extending min-\(p\) to other generative tasks, such as code generation or multimodal models, could reveal broader applicability and benefits across different domains.

\paragraph{Research into high-temperature regimes:} High-temperature regimes have been underexplored relative to low-temperature regimes. Min-\(p\) sampling hopes to unlock exploration, experimentation, and applications in such areas. Recent work in reinforcement learning has begun adopting min-\(p\) for trajectory data generation \citep{tajwar2025traininggenerallycuriousagent}, specifically citing its benefits for generating high-quality and diverse training data. In their Llama-3.1-8B-Instruct model implementation, they selected min-\(p\) sampling with temperature 1.5 and min-\(p\) parameter 0.3 after observing that this configuration consistently generated diverse training data while maintaining higher real-time accuracy compared to other sampling methods. This demonstrates that principles from min-\(p\) have valuable applications beyond text generation in areas like agent training and decision-making systems, where balancing exploration and coherence is crucial.

\paragraph{Human Evaluation Scope:} Our  involved participants selecting pre-generated outputs. We note that min-\(p\)'s popularity within the open source community for creative writing is interactive in nature; hence, we hope for adoption on interactive platforms such as the Chatbot Arena \citep{chiang2024chatbotarenaopenplatform}.

\paragraph{Combining uncertainty and CoT decoding methods:} \cite{wang2024chainofthoughtreasoningprompting} found a significant correlation between the confidence/certainly level of the final answer token choice and correct scores on GSM8K CoT, and that promoting diverse lower-probability token choices encouraged generating chains of thought that were beneficial for reasoning, resulting in higher scores overall.

This mirrors our hypothesis that choosing high-certainty tokens enables more accurate final answers, while diverse token choices benefit intermediate reasoning steps. For example, we note that GPQA scores on Mistral 7B increased from \(\tau=1.0\) to \(\tau=1.5\), but only with Temperature Only and min-\(p\). With the recent release of OpenAI's O1 models, which leverage CoT methods at inference for advanced reasoning capabilities, we note several novel decoding methods that combine uncertainty and CoT sampling approaches to improve model reasoning in a simple manner, with minimal added overhead and architectural changes \citep{wang2024chainofthoughtreasoningprompting, xjdr2024entropix}. We aim to actively explore such methods in future work.

\end{document}